\documentclass[10pt]{article}
\usepackage[explicit]{titlesec}
\usepackage{hyphenat}
\usepackage{ragged2e}
\RaggedRight

\usepackage{rotating}  
\usepackage{theorem}   
\usepackage{bm}  
\usepackage{amsmath,amsfonts,amssymb}
\usepackage{booktabs}  
\usepackage{pdfsync}   
\usepackage{graphicx}
\usepackage{latexsym}
\usepackage{amsmath}
\usepackage{amssymb}
\usepackage{fancyvrb}
\usepackage{subfigure}
\usepackage{colortbl}
\usepackage{enumerate}
\usepackage{pifont}
\usepackage{stmaryrd}
\usepackage{textcomp}
\usepackage{fncylab}
\usepackage{multirow}
\usepackage{paralist}
\usepackage{wrapfig}
\usepackage{longtable}
\usepackage[]{algorithm2e}
\usepackage[table]{xcolor}
\usepackage{lscape}
\usepackage{lipsum}

\usepackage{garamondx}
\usepackage[T1]{fontenc}
\usepackage{amsmath}
\usepackage{graphicx}
\usepackage{xcolor}

\usepackage{geometry}
\usepackage{fancyhdr}
\usepackage[labelfont={footnotesize,bf} , textfont=footnotesize]{caption}
\makeatletter
\renewcommand\tagform@[1]{\maketag@@@ {\ignorespaces {\footnotesize{\textbf{Equation}}} #1.\unskip \@@italiccorr }}
\makeatother
\titleformat{\section}
  {\normalfont}{\thesection}{1em}{\MakeUppercase{\textbf{#1}}}
\titlespacing\section{0pt}{0pt}{-10pt}
\titleformat{\subsection}
  {\normalfont}{\thesubsection}{1em}{\textit{#1}}
\titlespacing\subsection{0pt}{0pt}{-8pt}

\makeatletter
\newcommand\sixteen{\@setfontsize\sixteen{17pt}{6}}
\renewcommand{\maketitle}{\bgroup\setlength{\parindent}{0pt}
\begin{flushleft}
\sixteen\bfseries \@title
\medskip
\end{flushleft}
\textit{\@author}
\egroup}
\makeatother

\usepackage[sort&compress]{natbib}
\makeatletter
\renewcommand\@biblabel[1]{\textbf{#1.}\hfill}
\makeatother

\bibpunct{}{}{,~}{s}{,}{,}
\title{Towards Dynamic Feature Selection with Attention to Assist Banking Customers in Establishing a New Business}

\author{
Mohammad Amin Edrisi*$^{a}$ \\ \medskip
$^{a}$Macquarie University, Sydney, Australia \\  \medskip
mohammadamin.edrisi@hdr.mq.edu.au
}

\begin{document}

\vspace*{.01 in}
\maketitle
\vspace{.12 in}

\section*{abstract}

Establishing a new business may involve Knowledge acquisition in various areas, from personal to business and marketing sources. This task is challenging as it requires examining various data islands to uncover hidden patterns and unknown correlations such as purchasing behavior, consumer buying signals, and demographic and socioeconomic attributes of different locations. This paper introduces a novel framework for extracting and identifying important features from banking and non-banking data sources to address this challenge. We present an attention-based supervised feature selection approach to select important and relevant features which contribute most to the customer's query regarding establishing a new business. We report on the experiment conducted on an openly available dataset created from Kaggle and the UCI machine learning repositories.

\section*{keywords}
Business Process Management; Feature Engineering; Customer Success

\vspace{.12 in}


\section{introduction}
\label{chap:introduction}

Retail banks operate to make profits similar to other businesses, although money plays a role as inputs and outputs, unlike most companies. And because of that, operations in the banking industry mainly focus on risk management. However, to achieve higher revenue, re-engineering the current processes is playing a vital role.

In this study, the main goal is to create a win-win situation by proposing new services to small and medium-sized enterprises (SMEs).  The bank can adopt a new marketing strategy for customer retention and outstanding customer service to attract potential customers by offering a new product. This service will help the bank to prepare a statement of advice to find the optimum location based on requirements provided by the customer of the small business sector at the time of production. There is no estimate of inflation and future economic conditions.

The paper begins with a quick overview of businesses process in the banking industry as a \textit{System Research}. In particular, it looks at one of the opportunities that are the result of the intelligent revolution. As a start point, the importance of defining the understandable processes across multiple systems for stakeholders is described. There are several reasons for analyzing the process when the aim is dealing with process-centric queries. In a competitive market, enterprises are constantly searching for new value-added using the latent pattern in process analytics.

Following the discussion about business processes and their relatedness to big data and knowledge lake, \textbf{Section~\ref{Chapter 2}}, discusses the background, related work and State-of-the-Art.
After outlining the challenges we face and value proposition with new technological innovation, \textbf{Section~\ref{Chapter 3}}, briefs the reader about the applied methodology regarding data set and expected improvement using
Multi-task Learning (MTL) method, limitation of other methods and justification of MTL when tackling multi-criteria decision-making techniques for classification.
\textbf{Section~\ref{Chapter 4}}
examines different classifiers on the sample dataset to predict the success of new enterprise establishment. Analyse each classifier's performance and evaluate the methodology using evaluation metrics for the stated classification problem regarding the nature of data.
\textbf{Section~\ref{Chapter 5}} concludes the results and examines the future works.

\subsection{Competitiveness is More Complex in Fourth Industrial Revolution  }

Since 1960, the computing world has been evolving at a fast pace. The development of hardware and applications causes this progression of global reforming. Also, further innovation in decision-making is driven by mathematics. Recently, thanks to Artificial Intelligence (AI) technology and Machine Learning (ML) algorithms, the process of a more detailed and accurate analysis of big data reveals more insights into the cognitive revolution~\cite{CognitiveAug,cogFrankl}.

If the topic is placed under the microscope, then it becomes clear that more studies in this field trying to find brain patterns of accumulating knowledge and decision making process. For example, this trend is shown in books~\cite{B_Sapiens, B_21lessons} by Israeli historian Yuval Noah Harari, when he starts with humankind's history, highlighting cognitive revolution to the scientific revolution. He then has returned with 21 lessons on the importance of biotech and info-tech for the future of our life. As such, neurological scientist Robert Sapolsky has described our underlying behaviour elements, such as aggression, sympathy, and love. He has examined~\cite{B_Behave} one common factor in all of that behaviour, causality, and how we decide our reaction.

\begin{figure}[t]
	\begin{center}
	\includegraphics[width=1\linewidth]{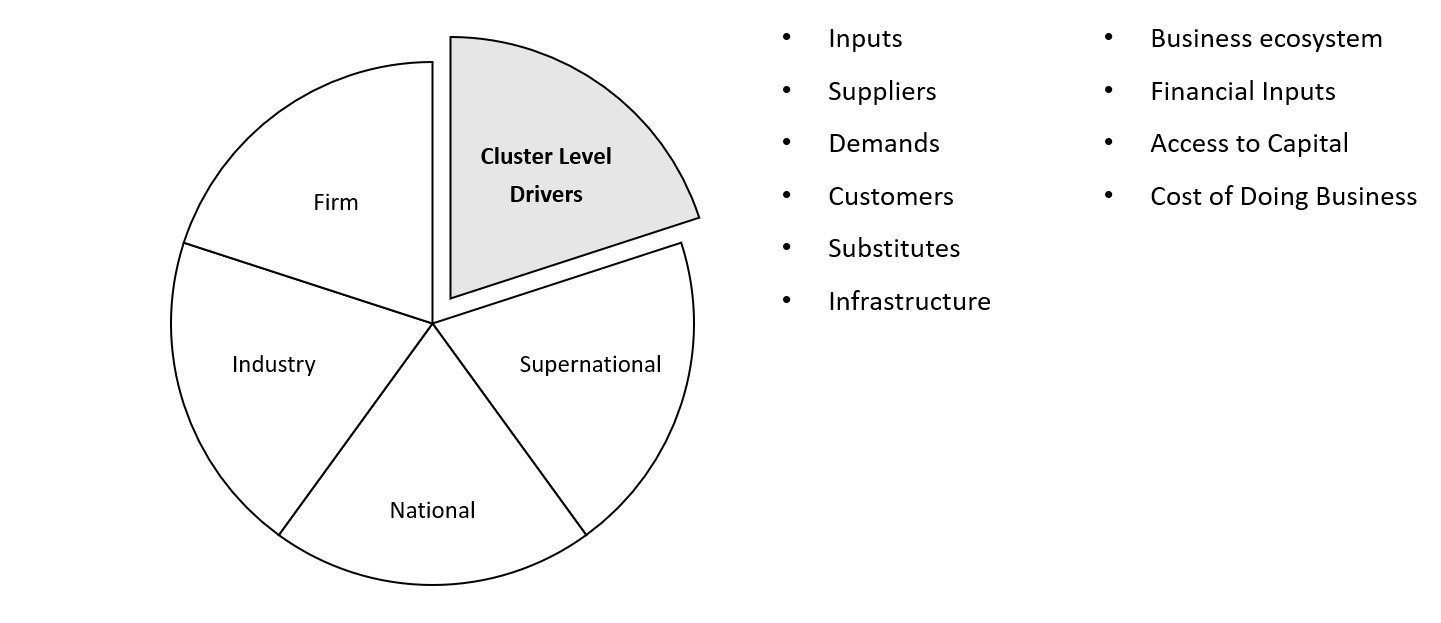}
	\end{center}
	\caption{Drivers level of Competitive Framework in Relation to  the Present Use Case, adopted from~\cite{Australiascompetitiveness}.
}
	\label{drivers level}
\end{figure}

Micheal J. Enright has developed a five-level competitiveness framework~\cite{Australiascompetitiveness} in the book, \textit{Australia's Competitiveness}. In his model, the driver of competitiveness has divided into five levels: the firm, the industry, the cluster, the national and the super-national level. The drivers related to global markets are out of countries' control, such as a pandemic situation or world war. Some macroeconomic features impact the whole market within a country, for instance, interest rate and government policies. Drivers on the micro-level are those that make one industry attractive for some business owners as well as creating barriers for others. The most prominent driver for the current study from the Figure~\ref{drivers level} is cluster-level drivers surrounding individual industries. The banking industry can provide valuable insights for new customers who want to establish a new small and medium-sized business employing a new data strategy that includes the local market's downstream drivers.

These days in a highly competitive market, the definition of competitive advantage is well-defined by computing powers, storing different types of data from a transactions perspective to even behavioural attribute of each customer, generating insight and making an informed decision.
Under these circumstances, we need a method to predict whether a newly established business can survive more than five years~\cite{SB_report,WA-SamllB} by examining the relationship between the current situations and the requirements of prospective business and features extracted from existing businesses.

\subsection{Problem Statement}

\begin{figure}[t]
	\begin{center}
	\includegraphics[width=1\linewidth]{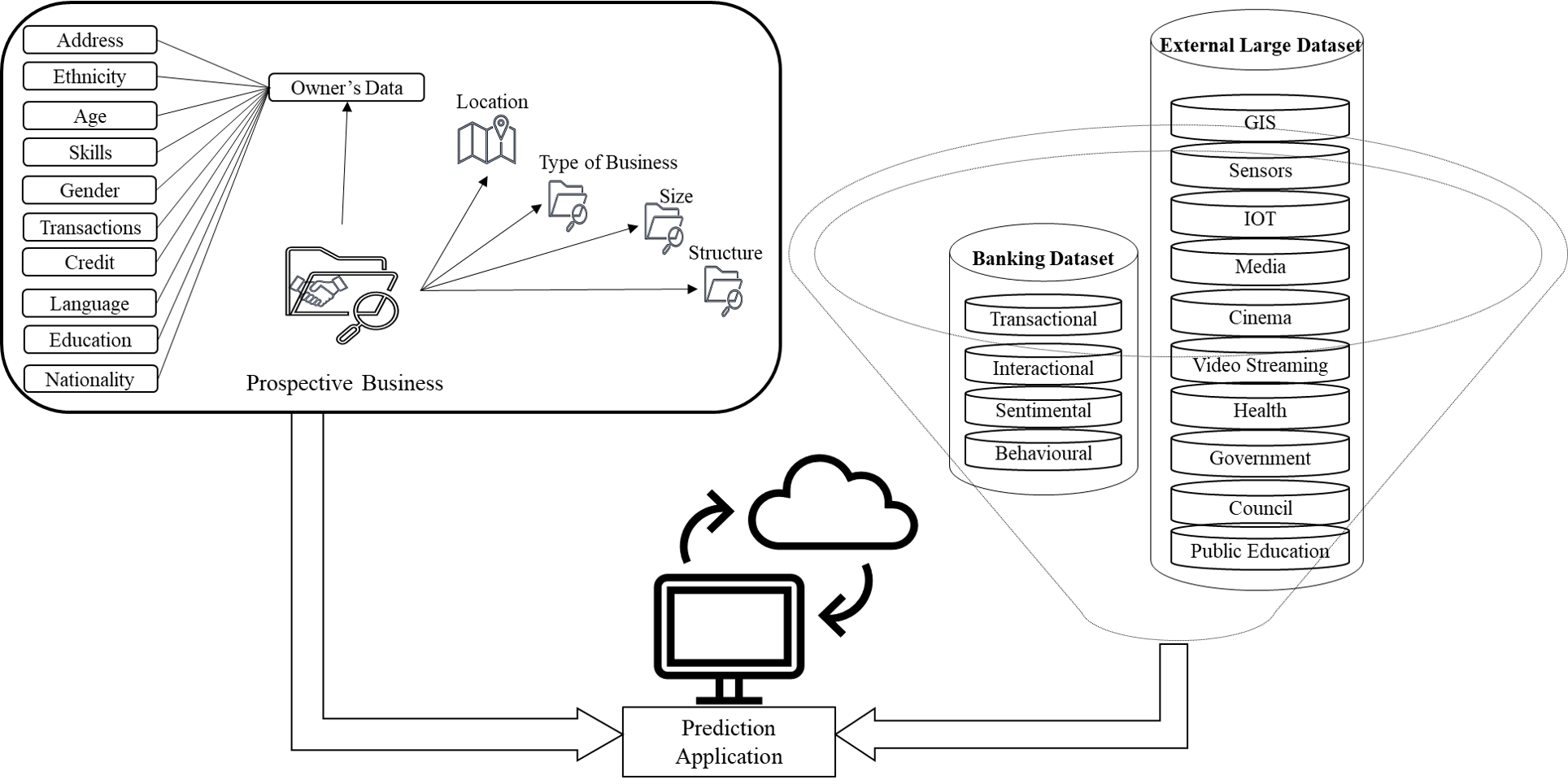}
	\end{center}
	\caption{Overview of AI-Enabled Prediction System in banking.}
	\label{AI-Enabled prediction system}
\end{figure}

Placement of successive business establishment can be viewed from different perspectives. There are stakeholders in every business, from business owners who want the most likely rewarding location to survive, make profit, and be successful to government agencies and financial institutions. As illustrated in Figure~\ref{AI-Enabled prediction system}, taking advantage of modern process analytics, ML algorithms, and big data will help market analysts to uncover successful criteria from raw data and support revenue-generating channel to a more significant profit.

From an economic perspective, banks are financing businesses and aim to mitigate their default risk. In addition to macroeconomic factors and legislation that can tighten lending practices, the physical store location can impact profitability~\cite{AIfuture}. The right site must be considered since, in the short-term, it can not be modified.
Looking more closely, it is apparent that banks can advise prospective owners to establish in the right place by making an intelligent choice using banking and external datasets. Creating this win-win situation for banks is feasible due to economies of scale. Gathering and analysing internal and external attributes are costly for individuals. Simultaneously, banks have to collect customer data per transaction, and the cost advantages of their operation scale enable them to provide this service.
As seen in Figure~\ref{AI-Enabled prediction system}, the two leading groups of data have been used to form a platform. There are cold data like transactional data that analysts can use for reporting queries and hot data like behavioural features using by businesses to respond to ad-hoc queries. To stay competitive, business stakeholders need to make an informed decision. As such, the future of banking goes hand in hand with cloud-computing investment, developing a data product~\cite{DataProduct} to ingest batch and stream data, computing over the data lake, and deploy analytics tool. The building blocks of the new data platform enable banks to advise their customers based on data-driven prediction.

\subsection{Motivating Scenario}

\begin{figure}[t]
	\begin{center}
	\includegraphics[width=1\linewidth]{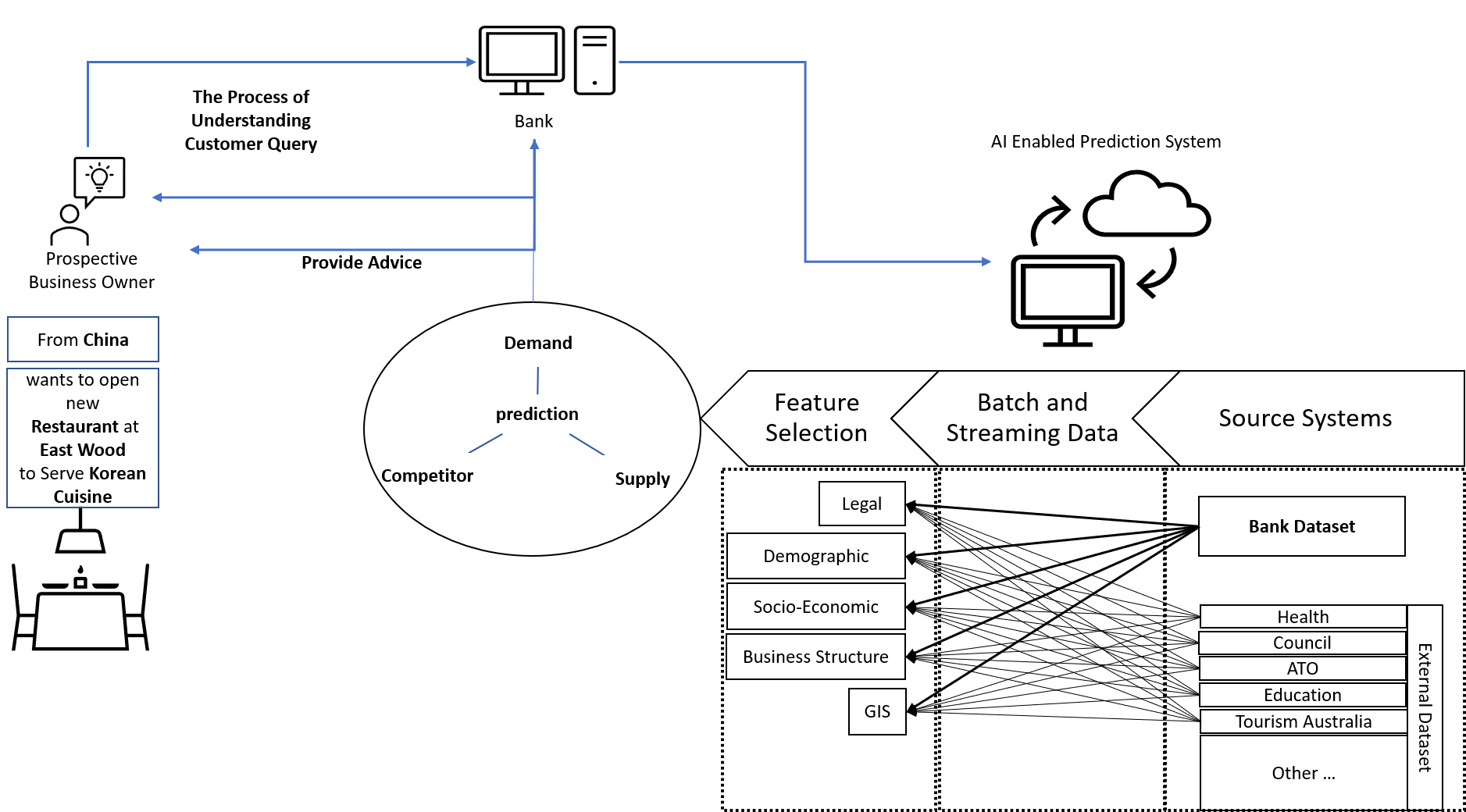}
	\end{center}
	\caption{Proposed Path to Provide Data-Driven Advice for Small Businesses.}
	\label{Prediction Model Motivating}
\end{figure}

Running a business is not like learning to drive a car by trial and error anymore. In this context, understanding the business environment is a key requirement to succeed. One of the most important aspects of a successful small business is the financial turnover, mainly from the demand side. This study will highlight those influential factors that can save time and money by analytical processing as a service from the banking industry and examine the impact of big data technologies on provide data-driven advice. As shown in Figure~\ref{Prediction Model Motivating}, this model will bring external data sources together with the transactional dataset to provide a holistic view of analysis. Transforming data to insight involves an intermediate processing step that can be both logically and technologically intensive.

The Australian tax office~\cite{SB_report} has defined \$10 million turnovers as a threshold for small businesses. Simultaneously, the same research suggests only 1.55\% of Australian enterprises have a turnover equal to or higher than this amount. Since the turnover of a business is subjective, we cannot oversimplify it. It is reasonable to consider another criterion to categorise small businesses, which is employment size. The combination of sole traders, micro and small businesses make almost 98\% of enterprises with fewer than 20 employees per business.

\subsection{Overview of Small and Medium-sized Enterprises (SMEs)}

\begin{figure}[t]
	\begin{center}
	\includegraphics[width=1.0\linewidth]{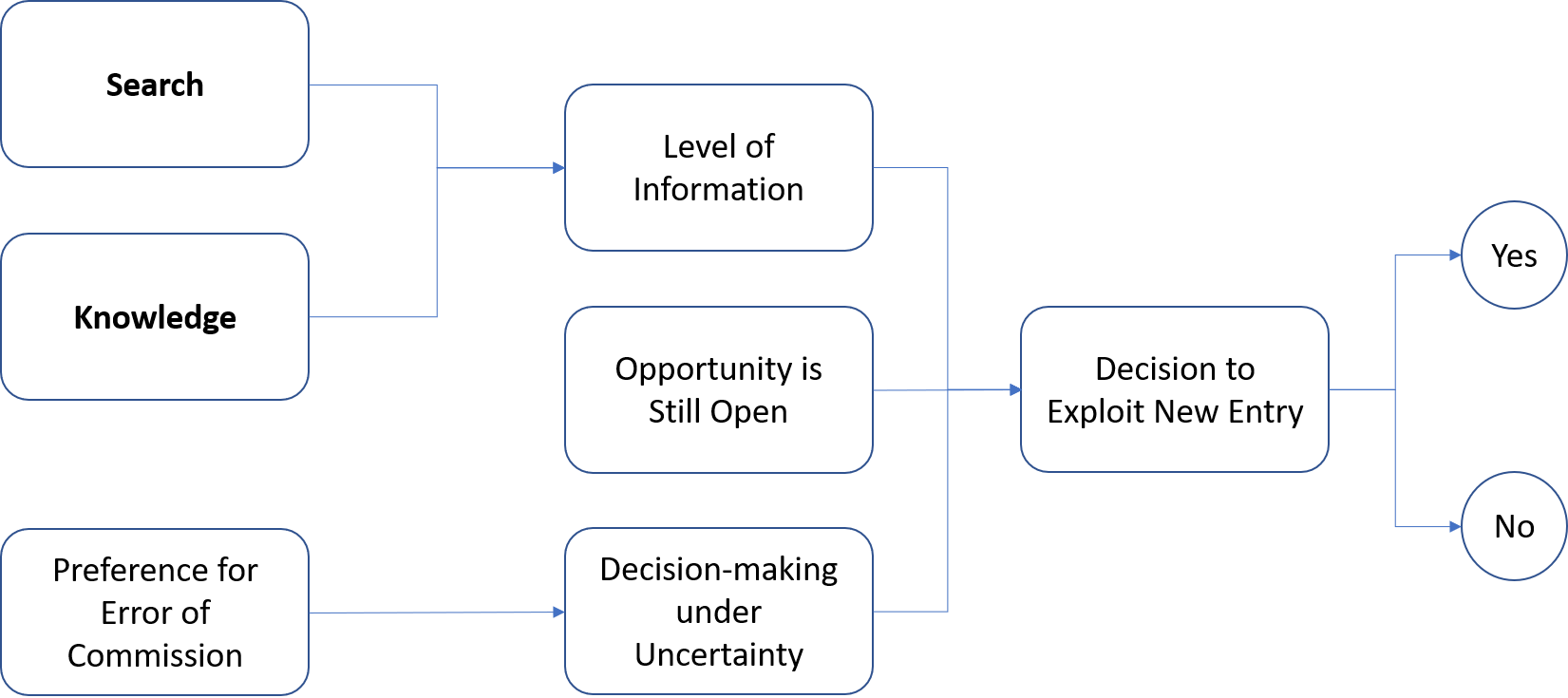}
	\end{center}
	\caption{Influencing Factors for Running a Small Business~\cite{2014IntegratingEdition}.}
	\label{Exploit}
\end{figure}

In this section, we will outline the potential factors~\cite{WA-SamllB} that have an impact on SMEs. It has been suggested that a small business's success or failure mostly depends on management skills, business plan, professional advice, working capital, and market research. The influential factors to start the new business is categorised in Figure~\ref{Exploit}.
Based on Figure~\ref{Exploit}, business owners can develop a set of questions and provide answers before establishing a new business.
As we can see, all three initial parts of this diagram are closely related to information and data accessibility. Since the business landscape has changed considerably in the past few years, SME owners need helps from professional institutions to tackle these issues. For example, the error of commission that arises from an overestimation of the internal or external situation can be prevented. Therefore, extract useful information needs queries that need to be analysed before establishing a new business, as shown in Table~\ref{table Business Checklist}. The last two columns demonstrate our ability to measure and data availability of the data.

Even though in most cases, management skills are subject to an individual, and out of control aspects like business and economic climate depend on domestic and international policies. Yet, a common strategy to tackle any problem is breaking down the components and, in this matter, dividing those influential factors into different categories and try to harness the data-driven one. Also, management skills had been affected by Management Information Systems (MIS) and technology enhancement. So, business owners need to seek professional advice. Similarly, we can see the same pattern for market research.

\setlength{\arrayrulewidth}{0.2mm}
\setlength{\tabcolsep}{18pt}
\renewcommand{\arraystretch}{1}
 \begin{table}[h!]
\caption{Considerations to Move into SMEs Market. }\label{table Business Checklist}
\centering
    \begin{tabular}{ |p{2cm}|p{4cm}|p{1.5cm}|p{1.5cm}|  }
        \hline
Category & Attributes & Measurable & Data Availability   \\
        \hline
Personal Assessment & skills & No & No \\
                    & business understanding & No & No \\
                    & management experience & No & No \\
                    & personalised financial plan  & Yes & Yes \\
Business Structure  & legal structure & Yes & Yes \\
Business Planning & skills & Yes & No \\
Break even & cost & Yes & No \\
 & finance & Yes & No \\
  & tax & Yes & No \\
Location & accessibility & Yes & Yes \\
         & visibility & Yes & Yes \\                                  & people & Yes & Yes \\
         & premises & Yes & Yes \\
Employment & employees & Yes & No \\
Market Consideration  & strategy & No & No \\
              & promotion & Yes & No \\
             & advertisement & Yes & No \\

       \hline
    \end{tabular}
\end{table}

\subsection{Contribution}

Establishing a new business may involve Knowledge acquisition in various areas, from personal to business and marketing sources. This task is challenging as it requires examining various data islands to uncover hidden patterns and unknown correlations such as purchasing behaviour, consumer buying signals, and demographic and socioeconomic attributes of different locations.
While there have been various studies of site selection~\cite{location-problemStatement,LocationRestaurant}, for small and medium-sized businesses and different use cases~\cite{AIfuture} of big-data applications in the banking industry, the implementation of big data analytics in practice with the goal to enhance SMEs productivity in the banking domain is unclear. There are two groups of contribution as follow:

\begin{itemize}
    \item Better risk management by operational efficiency improvement of small businesses.
    \item Train a single model to predict several criteria of a successful business.
\end{itemize}

In this paper, the main contribution is proposing a new practical implementation of AI technologies on top of the big data platform in the banking industry that can improve risk management, customer acquisition and generating a new revenue stream. Banks are still the main channel of financing small businesses. Although considering the default risk is critical for banks in credit allocation, they can help their customers make an informed decision and mitigate the risk. Studies show the dominant pattern in banking data mining methods is tackling risk management and customer relationship management problems~\cite{Generalsystematicliterature}. Therefore, helping with the issue of end-to-end process methodology is examined in this paper. Unlike other studies in the area of efficiency improvement that have considered a concrete banking business problem~\cite{transformingparadigm}, the proposed solution will create a link between bank risk management, customer retention and their success to establish a profitable business.

On the other hand, AI-enabled customer solution provides advice to the customers to establish their new businesses in the most likely profitable location. Humans usually acquire their knowledge by connecting outcomes from their past experiences and applying that wisdom to facilitate the learning process of new tasks. Accordingly, the checklist for business starter may contain close to 20 criteria. By applying a Multi-Task Learning algorithm, we can take each criterion as a separate task and predict the outcome with lesser-focus on a single task, such as buying-driver, sales forces, customer's bargaining power, competitors and supplier power. In this model, optimising one loss function helps optimisation the process of the related loss function for another task. The main contributions of this paper include:
(i)~a novel framework for extracting and identifying important features from banking and non-banking data sources to address this challenge; and
(ii)~an attention-based supervised feature selection approach to select important and relevant features which contribute most to the customer's query regarding establishing a new business. We report on the experiment conducted on an openly available dataset created from Kaggle and the UCI machine learning repositories.


\section{Backgrounds and State-of-the-Art}\label{Chapter 2}

Intelligent decision making using big data can create a competitive advantage for any organisation. In particular, it is vital to realise the importance of an intelligent revolution. There is no doubt about the importance of algorithms to harness data and create useful insights from start-up companies to big multi-national companies. Still, the question is about forming an applicable model so that any stakeholder from the end-user to the top management can appreciate the benefit and understand the value-added.

The banking industry has organized customer's data for many years, mostly structural data and in a tabular format. Therefore, once the computation ability of machines enabled them to use those data, they have tried to understand the raw data sets. There is still room to improve from unstructured data analysis for banking functions such as an identification process without the necessity of ID card to the up-selling by recognising customer needs via the chat-bots.

In this section, we look at the business processes in general and banking procedures in particular. Then, based on identified processes, we will focus on data-driven and knowledge-intensive cases. After that, we will specify the importance of customer-related activities in retail banking and designing some win-win situations for both parties through the AI-enabled banking system. Finally, we indicate the importance of big data and methods already used in different banking sectors.

\subsection{Business Processes}

To support enterprises to run effectively, the correct data at the appropriate time will enable the right people to use the right process. In this section, we focus on the business process and the likelihood to discover actionable information from the large data-set. At an organisational level, a business process delivers targeted results with the goal of support of strategic decisions with understanding and analysing them. Business analysts usually talk to stakeholders to collect and analyse the area of efficiency improvement~\cite{B-BP-IBM}.

Definition of business activities indicates there are three main categories of tasks that should be done simultaneously to run a profitable business~\cite{process1,process2,process3,process4,process5,process6}. The structured activities of operation, marketing and finance need a method to align the human, tools and machine to get the job done. From a systematic point of view, we call this description as a business process. Data and AI, in addition to modern workplaces and business applications over the cloud, helped organisations to achieve a greater degree of sustainable business development. However, this degree of improvement has an impact on conventional structured process management by improving communications and eliminate barriers between ultimate users and those who make the decisions. The next step is to connect the intangible components of the unstructured process to quantifiable process knowledge~\cite{B-processanalytics-Beheshti}.

Not only businesses but also government agencies and not-for-profit organisations have to define processes once there is an interaction between different parties~\cite{B-processanalytics-Beheshti}. Since the computing power of new technologies enables enterprises to collect, store and analyse data on a large scale, the business process strategies have to practice modern techniques if they want to survive. In this study, we consider those systematic BPs that are convertible to an automated and thus process improvement in the business context is related to them. Before diving deeper, let's first distinguish between analysis and analytics that sometimes use interchangeably.

\subsubsection{Changing the Analysis of Businesses}

From the information system points of view~\cite{MIS,2014IntegratingEdition}, there is a process data pipeline between different activities. By breaking down the entire dataset, examine the relationship among them and answering questions like how and why, we will form the analysis. On the other hand, business analytics generally refers to the future instead of explaining past events. It explores potential future ones, and it involves logical and computational reasoning for pattern recognition and exploring what you can do with them in the future. Qualitative analytics is a branch of analytics that applying insight and expertise in combination with the analysis to plan the next business move. In contrast, in quantitative analytics, the main concern is formulas and models in conjunction with numbers have been gathered from the analysis.

For example, suppose a business analyst has a great understanding of customers' requirements. She has performed a very detailed analysis of home loan types and feels sure about which target market to follow. Also, may use this intuition as qualitative analytics to decide what kind of services to start selling to which group. But relying on past sales and user experience data, she will make an informed decision for when to introduce the new service to the same group. She could predict which stage of the customer journey would be best to do that, even though she is not a salesperson.

\subsubsection{Every business process should create value}

Ability to adopt new process transformation after the COVID-19 pandemic is a strong motivation for businesses with the conventional process to move to the new one while ensuring that they are meeting their requirements.
First, we will have a look at three dimensions of the processes that help us to understand the characteristics of the business process.
 Each dimension has its classifications. For example, \textit{process paradigm} refers to structured, semi-structured and non-structured processes. In each category, involvements of human skills determine the class~\cite{B-processanalytics-Beheshti}.

One of the process paradigms is the AI customer retention process that searches critical behaviour, identifies the effective conversation with the right person at the right time for customers you should be engaging with. It's hard to determine which customer churn every year and will be out the door soon. More importantly, why is that customer going to go~\cite{CustomerValueGeneration}. So banks can intervene to stop customers by making the predefined schema-based approach to understand customers and how they feel about enterprises.

Designers can employ the flow of activity to activity for the structured process to represent the desired model. The next tactic is suitable for the process with few constraints so few rules would be reasonable, and the third one that is artifact-centric approach talks about the relationship between business artifacts and their diverse requirements. In conjunction with application-layer characteristics, implementation technologies reflect the third dimension from the workflow engine, which introduces the process instance to the rule engines that smart to match the pattern of facts for a process and generate rules. And finally, the less desired method is a hard-coded program to execute a process.

A common observation in the literature suggests there is a lack of end-to-end process support and thus the light of all available data, a suitable solution would include a bridge between the structured and unstructured processes.
So, to summarise the argument in this section, for the banking industry we have to perform several things: first, define who performs the work in the process paradigm in either \textit{human-centric process} or \textit{system-centric process}. Then, apply Attribute-based routing to cover both pull or push systems. Finally, optimize a model powered by serious AI muscle~\cite{B-BP-IBM, B-processanalytics-Beheshti}.  following this introduction, we will look at the data-related BPs and finally talk about rooms to improve for retail banking as one of the leading industries that collect and analyse a substantial amount of data every day.

\subsection{Business Process in Retail Banking}

\subsubsection{Conventional versus Modern Imperatives in Banking}

The reinvention of BPs is a vital move in the highly competitive market of economic volatility, market uncertainty, regulatory changes and customer needs. Regardless of industry, optimising the process, better compliance controls and providing outstanding customer services would be the three critical areas of consideration. For each field, research shows~\cite{B-processanalytics-Beheshti,AIfuture} both established processes and the possibility of generating processes by users.

One interpretation of the redefinition business process can be borrowed from programming blocks and their relationship. The sequence is the first block and is the list of everything that needs to be done in an orderly manner or parallel.  Selection is the second building block and identifies how to do things and if statements are the most common example in programming languages. The third one is looping, or iteration consists of repeating sequence and selection activities several times. It is interesting to bring the same concept to business processes based on the business or customer requirement. The difference here is an iteration between business users and front-line workers who update process knowledge.

According to the \textit{Cambridge Centre for Alternative Finance}~\cite{transformingparadigm}, FinTechs statistically adopted AI in process re-engineering and automation domain more than risk management, client acquisition, and customer service. So this can show the gradual movements from conventional automation towards AI-enabled process analytics.

The future world of business is uncertain and complex; therefore, the role of AI applications in the decision-making process is inevitable. On the other hand like any capital investment, successful AI adoption heavily depends on trust and financial return. Considerations would be the precise definition of the financial statement's specific outcomes and justifying the process implementation.
Most organizations consider AI-based systems as a black box with no transparency. However, research related to AI development~\cite{BankingAnalysisML} through machine learning has mentioned three structures with advantages that save money and time. Table~\ref{table ML Structure}, has summarised the efficiency improvement in the banking sector using those structures.

\setlength{\arrayrulewidth}{0.2mm}
\setlength{\tabcolsep}{18pt}
\renewcommand{\arraystretch}{1}
\begin{table}[!h]
\caption{Useful Machine Learning Structures~\cite{BankingAnalysisML,MUSTRead} in Banking Industry.}\label{table ML Structure}
\centering
\begin{tabular}{ |p{2cm}|p{4cm}|p{5cm}|  }
\hline
Structure& Application& Advantages   \\
\hline
Reinforcement Learning & Fraud Detection & Containing multiple agents in a real world, so the learning paradigm is sequential\\
 & Recommender Systems & Relying on user cognitive aspects~\cite{Beheshti2020}  \\
 & Conversational Chat-bots & \\

Supervised Learning & Customer Behaviour Modelling & Patterns of electronic transactions based on demographic and behavioural features \\
 & Analysis  of  credit  scoring & For the last ten years random forest and extreme gradient boosting have demonstrated the better job~\cite{creditscoring1,creditscoring2}\\

 & Customer Relationship Management (CRM) & Predicting customers who are likely to churn~\cite{SVMchurn}  \\
 &  & Customer loyalty and retention  \\

Unsupervised Learning  & Customer Segmentation & K-Means for Clustering~\cite{segmentationK-means} \\
 \hline
\end{tabular}
\end{table}

\subsection{Data Driven Processes}

Change the value creation using data generated by smart products has already introduced a business model innovation for companies such as Uber. This change of converting business requirements to the business model will deliver the systematic value proposition for existing machine learning processes~\cite{DataSynapse,Alir1,Alir2}. This progress has helped organisations to transform data into knowledge. A study by Alina Sorescu~\cite{DDmodel}shows the importance of business models leveraging external data as well internal data for companies to create a core competency in the market. Adopting this strategy has helped them to move from microanalytics to macroanalytics.

According to Dr W. Edward Deming who contributed to Japan's industrial rebirth and worldwide success :
"Without data you're just another person with an opinion." Since IT-Driven restructures from the 1980s, many companies collect data and employ those data to create value. But when we talk about a Data-Driven organisation, we mean the organisational \textit{culture} that acts based on data, \textit{technology}, and \textit{abilities}~\cite{B-DD}.

Different components are needed to transform a business into a data-driven one. The process starts with data collection of \textit{reliable} data as an ingredient. Literature shows data quality management~\cite{DD-qualityconsideration} is crucial since any robust model needs accurate data.

Once those accurate data are collected, the next step is about \textit{querying}, \textit{joining} and \textit{sharing} those data. From this step, the essence of \textit{human} skills is critical since the machine does not know what would be the right question to ask from data.
Therefore, aligning algorithm and systems, defining metrics and delivering value in aggregation with business context are still human tasks or is a human-centric~\cite{DD-Human-Centered-ML,B-DD} process.
The rest of the transformation to be a data-driven organisation is about migration from \textit{reporting} to \textit{analysing}. Because to manage a system, we need to provide a causal explanation of what is the reason behind some change. At this stage, we can see the collaboration of machine and human skills. Based on a study by R. Salvador et al., five biological factor~\cite{DD-gutfeeling} impact the decision-making process, and the process of transforming into a data-driven business will be completed if those extracted insights can \textit{influence} decision-makers.

\subsection{Knowledge Intensive Processes}

The activity execution order in business process modelling focuses on a significant relationship within the control flow. However, knowledge workers, such as a business analyst or a company manager, can identify other relationships~\cite{CoreKG,Alir3,Alir4,Alir5}. Also, works with large amounts of data need to extract the appropriate data and carrying out analysis to solve business analytical problems. This journey of applying solutions starts with a piece of code to retrieve data from the database. However, the next event prediction is not an easy task~\cite{BP-Beyondarrow}.

\begin{figure}[t]
        	\begin{center}
        	\includegraphics[width=1\linewidth]{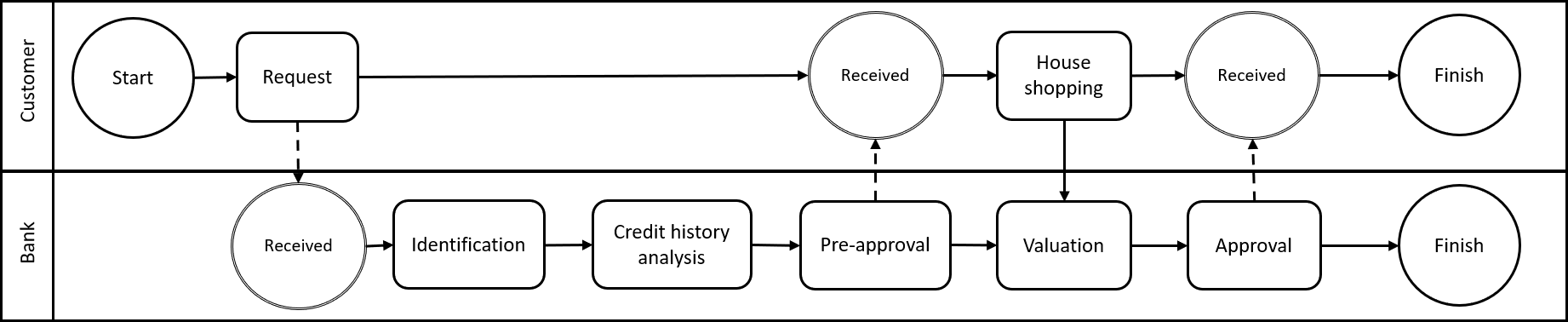}
        	\end{center}
        	\caption{A Simple Home Loan Process.}
	        \label{Home Loan Business process}
        \end{figure}

In the case of multiple kinds of relationships of interest among such activities, there are activities shown by the control arcs in a simple home loan process model of Figure~\ref{Home Loan Business process}.

As an example, the intrinsic execution order is assumed to approve a home loan while each step in the real world influences the real decision-making process. So, for more informative explanatory analysis, there should be a knowledge-intensive process to tackle real-world constraints. As a second example, let's consider demographic characteristics as an essential factor for valuation from a business analyst point of view. This example reflects the current procedure of a simple domain which is less challenging to understand by human and could easily be identified by the few line of code if the analyst has adequate knowledge of a particular specialization. But, how about an organisation doing day-to-day transactions with thousands of employees and millions of customers.
In this case, real-time adjustments have an effect on underlying domain constraints and can take into consideration using AI-enabled banking prediction.

\subsection{Big Data}

Even though there is a debate on market control by using consumer data, M. Nuccio et al., have focused on the widespread availability of big data to major players in the market, such as social media platforms~\cite{BigdataHellorheaven-4M}. They evaluated the possibility of market power abuse by major players given a large amount of data available. They concluded that major players having access to big data does not necessarily lead to abuse in the digital marketplace. Still, it also leads to more innovation by those who own the data.
Since 1990 companies have widely accepted the five forces framework developed by Micheal E.Porter to analyse the competitive environment. This model is particularly famous for new starter businesses regarding drivers' strengths to determine the potential profit. This framework allows a company to assess collective forces from the buyer side, supplier side, market attractiveness and substitute products. It is time even for SMEs to identify the drivers of competitiveness if they want to survive in the fourth industrial revolution~\cite{PorterDiamondModel}. The big data terminology involves the variety and volume of the generated data and enhanced the potential value of creation values. For example, companies might have a good product view of a customer, but they do not have a single customer view. Big data technology facilitates the creation value and profits from the influential attributes of successful small businesses.

In recent years much work has been done to highlight the importance of big data on business and economic growth. The literature in this area supports the notion that big data is becoming increasingly critical for the success of entrepreneurs, improving productivity and enhancing customer experience. In one study, the importance of utilising big data derived from social medial platforms to formulate marketing strategy~\cite{BigData-1M} was explored by P. Ducange et al., highlighted the various dimensions available in social big data that can assist marketers to make fast decisions to enhance customer experience. They concluded that given the complexity and size of information available on social media, there can not be a single platform to guide the markets in their strategy setting and decision making.

In another study, A. Popovič et al., examined  the impact of Big Data Analytics on high-value business performance. Their investigation focused on the impact of big data on operations management of the manufacturing sector~\cite{impactofbigdata-2M}. They assessed the impact of big data on manufacturing companies with different levels of data usage maturity. They concluded that to improve performance the manufacturing firms needed to have the capability and good data strategy, management support, investment, and employee engagement.

However, due to drastic changes in applications and technology, big data analysis is an alternative for designing a questionnaire to gather the determinant features that contribute to creating a competitive advantage~\cite{nabnab1}. F. Ciampi et al., have looked at the relationships between Big Data Capabilities and Business Model Innovation~\cite{impactofbigdata-3M}. Their research confirmed that businesses could have a more entrepreneurial orientation regarding their strategy with Big Data Analytics Capabilities. The availability of big data resources enables better risk-taking and an innovative approach to decision making. It can act as an enabler of success for the implementation of new ideas.

While much of the recent work in the area has concentrated on process mining for available event logs to support daily operations, the introduction of big data has changed a commonly understood process flow from process discovery to process improvement. Yet, the process mining use cases have not found the solution for to-be processes. For example, for real-world use cases, continuous changes have been made to data velocity as the main focus in this discipline~\cite{BP-BD}. As for variety, data sources are more than historical DBMS and involves images, audio and video.

\subsubsection{Data Lakes}

As we persist in applying ML techniques in the era of big data, we see another hand in hand terminology called "Data Lake". Data production in the last two decades is not limited to structured data anymore. Data generated by social media, the Internet of Things and even historical transactional data warehouses require a new architecture, management and analysis system~\cite{DLarchitecture}.
Academic literature on data lakes that addresses conceptual organising data identifies two overall architectures, Zone and Pond. According to the degree of processing data takes, the zone will be assigned. For example, no processed data will be categorised as zone zero. From the pond architecture of view, data flow is defined based on its characteristics, starting from raw data pond to archival data pond~\cite{DLchallenge}.

Even though we have mentioned the basic characteristics of big data like variety and volume, there are more considerations at an enterprise level that leverage big data. Enterprises previously could extract meaningful insights from transaction data by transforming those data to analytics data. However, today we need a new structure for enterprise data management. Internally generated data have their importance for different end-users, but issues around technologies to store and retrieve external sources play a crucial role in a data-driven market~\cite{B-DataLAke}.

While most enterprises used to have their data in a data warehouse, at least in three aspects, there is a difference between a warehouse and a data lake. First one is capacity; the regulatory definition for warehouses makes it hard to collect data as much as needed. The second characteristic is the Schema-on-Road approach which requires a schema for extraction, transformation and loading tasks. The last one is related to ML models since data can be unstructured and come in multiple forms, so data scientists as users can easily applying predictive modelling techniques on top of data lake which was is not possible with old data warehouse architectures~\cite{DL-BF,DL-2017}.

\subsubsection{Knowledge Lakes}

Centralised raw data islands to date lake to store and help analyst to curate them later is a one way to contextualised data anytime~\cite{Beheshti-BP-IKL}. Knowledge mining is an emerging category in AI to solve how to act on your content. There is a growing problem for a few different reasons; one is just about generating data and massive data sources outside the organisation. But the bottleneck here is that a lot of this data is unstructured. For example, analysing PDF documents is difficult to manage, and we need to use the unseen information intelligently and tackle different issues.
It is almost easy to develop a single model over one data source to solve one specific technical problem but knowledge mining providing knowledge workers with a deep sort of content understanding that is a more considerable challenge. This deep content understanding that business analyst can use starts with ingesting all types of data types such as unstructured ones and applying a combination of AI services together. Use more than one technique to extract insights from all sources that you might have. Once you have this deeper understanding, reveals valuable underlying information in all content depending on the end-use case.

\subsection{Banking and Customer Success}

There are several aspects to the problem to be examined, particularly ML and AI's different banking domain purposes~\cite{Generalsystematicliterature} regarding data-driven business processes and adapted banking industry practices. This part focuses on the customer as shown in Table~\ref{table literature}, including personalised procedure from customer identification to cross-selling.
The banking industry operates to make profits similar to other businesses regarding three main functions like marketing, operations, and finance. Although, unlike most businesses, money plays a vital role in both inputs and outputs of their procedure. so, we have to consider the functions of money and those attributes to understand the business and harness the prospective problems. From the economic perspective, money is the medium of exchange, store of value, and unit of account. Each one of those roles is evolved during the time from agriculture age to the information age and nowadays we have the privilege to take advantage of computing powers, storing different types of data and generating insight in the artificial age. In addition to that, the financial industry is highly regulated and defines new products and services that need careful consideration by R\&D. Putting together those facts take us to a greater challenge to make competitive advantages based on core competencies in a highly competitive market~\cite{CustomerValueGeneration}.

Today's ability to store huge amounts of data from a transactional perspective to even behavioural attributes of each customer as well as computing capacity enables us to look from different angles and be a winner in this market. Generally, operations in the banking industry mainly focus on risk management, customer service improvement, marketing strategies and customer retention. But, to achieve higher revenue, informed decision making in production and process re-engineering is playing a vital role~\cite{Chiorazzo2018}. In this study, the main concern is about advancing revenue generated by small and medium-sized businesses by providing them optimum positioning strategy and bring the opportunity to them to make a greater profit. 

\setlength{\arrayrulewidth}{0.2mm}
\setlength{\tabcolsep}{18pt}
\renewcommand{\arraystretch}{1}
 \begin{table}[h!]
\caption{Categories of data mining application in the banking domain }\label{table literature}
\centering
\begin{tabular}{ |p{2cm}|p{4cm}|p{5cm}|  }

\hline
Main Purposes& Field& Example  \\
\hline
Risk-Oriented & Risk Management Processes & Credit risk scoring~\cite{CreditRiskscoring,creditscoring1,creditscoring2} and default prediction \\
 &  & Prediction of financial distress \\
 &  & Credit decisions for private and corporate customers  \\
 & Identification and Prevention of Fraud Behaviour & Anti-money Laundering (AML)~\cite{AML}   \\
 & Overall Economic Predictions & Market risk~\cite{MArketriskmgmt} and asset management    \\
Customer-Oriented & Customer Behaviour Modelling & Patterns of electronic transactions based on demographic and behavioural features \\
 & Customer Relationship Management (CRM) & Predicting customers who are likely to churn~\cite{churnprediction}  \\
 &  & Customer loyalty and retention  \\
 &  & Customer segmentation  \\
 &  & Customer value identification  \\
 &  & Customer targeting in sales campaigns  \\
Efficiency Aspects of Bank's Infrastructure  & Automated Teller Machines (ATMs) and branch networks & branch location~\cite{branchlocation}  \\
 \hline
\end{tabular}
\end{table}

From the probability of default (PD) model to a fuzzy logic model for credit decision making, researchers have considered objective and subjective attributes to propose a mitigation way for risk management in financial institutions. Statistical tools to carry out risk analysis are not equipped enough to tackle the massive data generation volume. Also, multi-dimensional attributes of customers make more challenges regarding the multi-criteria decision-making process~\cite{CR-Multi}. The review has shown that the application of machine learning in the management of banking risks such as credit risk, market risk, operational risk and liquidity risk has been explored; however, it does not appear adequate with the current industry level of focus on both risk management and machine learning. A large number of areas remain in bank risk management that could significantly benefit from the study of how machine learning can be applied to address specific problems.	

Demographic and behavioural identification of customers has impacted on developing predictive models in the banking industry. Consequently, analysing internet banking users has justified the cost of developing new analytical methods and infrastructures. By identifying the behavioural  attributes of current customers, process improvement is possible to provide better services for them and adopt strategies for acquiring potential non-users~\cite{internetbankinguser}.

Companies should analyse the collected data effectively to determine strategic road maps compatible with business processes. The size of the Bank and operational data make classical productivity measurement methods impractical. Many organisations have not started to process their data to obtain practical and useful information regarding customer qualities and customers' buying patterns. Being successful and sustaining this success in the competitive market for the powerful institutions in terms of raw data, however insufficient in terms of qualified knowledge. Regarding the evolved role of internet banking, a significant portion of banking services and products are offered through branch channels, and branch distribution network optimisation has a critical proposition in terms of profitability and efficiency.

\section{Conventional Site Selection}

        \begin{figure}[t]
        	\begin{center}
        	\includegraphics[width=1\linewidth]{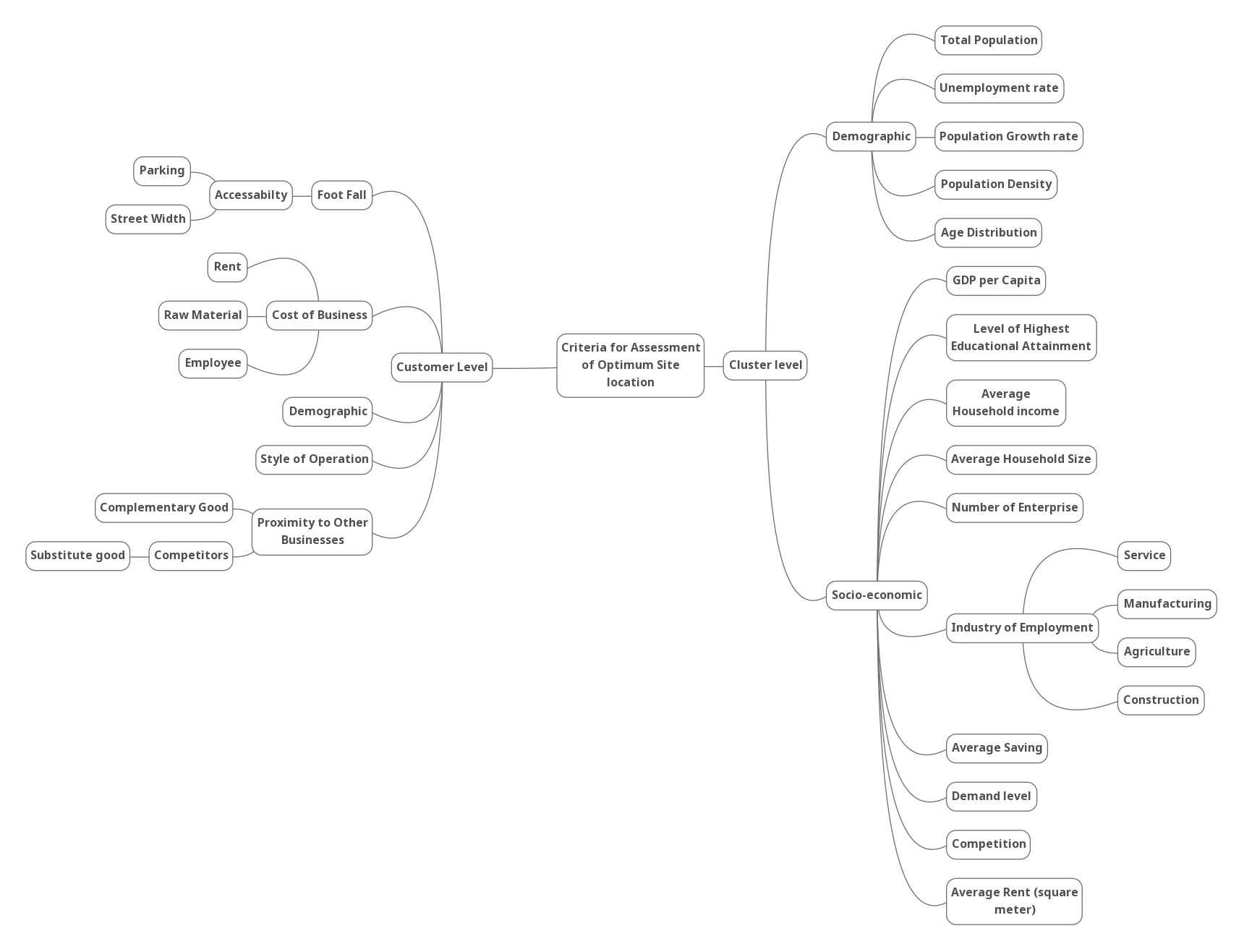}
        	\end{center}
        	\caption{Criteria for Site Selection.}
	        \label{Criteria}
        \end{figure}

Previous studies have identified criteria to invest in strategic retail location selection. As Yildiz et al.~\cite{SOTA_retaillocation} states, socio-economic factors that contribute to this investment decision include demographics, competitors, proximity to other businesses, and even average saving accounts per person. To solve a multi-criteria problem, they obtained weights using an Analytic Hierarchy Process and then ranked them based on the expert's opinion. Finally, it employed the Monte Carlo simulation to propose applicability. This process runs three different algorithms sequentially on alternative cities, final result is based on human judgments, and done on a cluster level no customer level. Figure~\ref{Criteria} illustrates the overview of factors that had been used to select an optimum site for retail businesses.

Traditionally, with the scarcity of data, a pair comparison analysis between factors could help us choose the best site. Conventionally, we could identify and rank the affecting criteria and sub-criteria based on the domain experts. At the same time, we want to predict based on historical data generated by the wide range of suppliers, customers, GIS and even behavioural attributes~\cite{demographicattributesfrompurchasedata}. From the demographic point of view, harnessing the big data based on customers' purchasing behaviour will enable us to do better basket analysis. Those personal attributes are assumed as accessible.
Consequently, it would seem an intelligent response to the problem is needed at both feature selection and weighting those features. therefore, applying AI technologies is a response to changing circumstances.

\subsection{AI-enabled Banking Methods - State of the Art}

Improving multi-object detection methods ~\cite{SOTA-multiobjectdetection} in deep learning is challenging since many applications, from self-driving cars to anomaly detection, are still looking for the right architectures. As such, since most cases involving multi-criteria decision-making, the definition of the final task and the best-shared representation between tasks have led researchers to find the optimal solution for generic and specific object detection.

Previous sections covered the existing related work of significant business success factors, data-driven processes, and big data applications in the banking industry. Also, discussed ways of the value proposition by creating a win-win situation for banks and their customers. As such, in that section, we have explained the importance of multi-criteria decision making. This section will introduce the ML field's recent algorithms to deal with multi-task decision-making to adjust the most relevant latent to predict small business success.
Even though we are trying to predict one out of two possible labels from collected data, the prediction of a successful business is not a binary classification anymore. For example, we can not easily have the type of business, store size, and the popularity of the area, and then predict whether a person will turn to a customer or no. The true definition of the process, product, customers, and suppliers will help us to extract inputs. For each business's primary function, we need to understand and predict the number of leads convertible to sales. So, to find out the prospective customer, we will break down the attributes of the existing customer~\cite{CustomerTargetingNeuralN,hamidNNet} from established businesses and extract the features.

\begin{figure}[t]
        	\begin{center}
        	\includegraphics[width=1\linewidth]{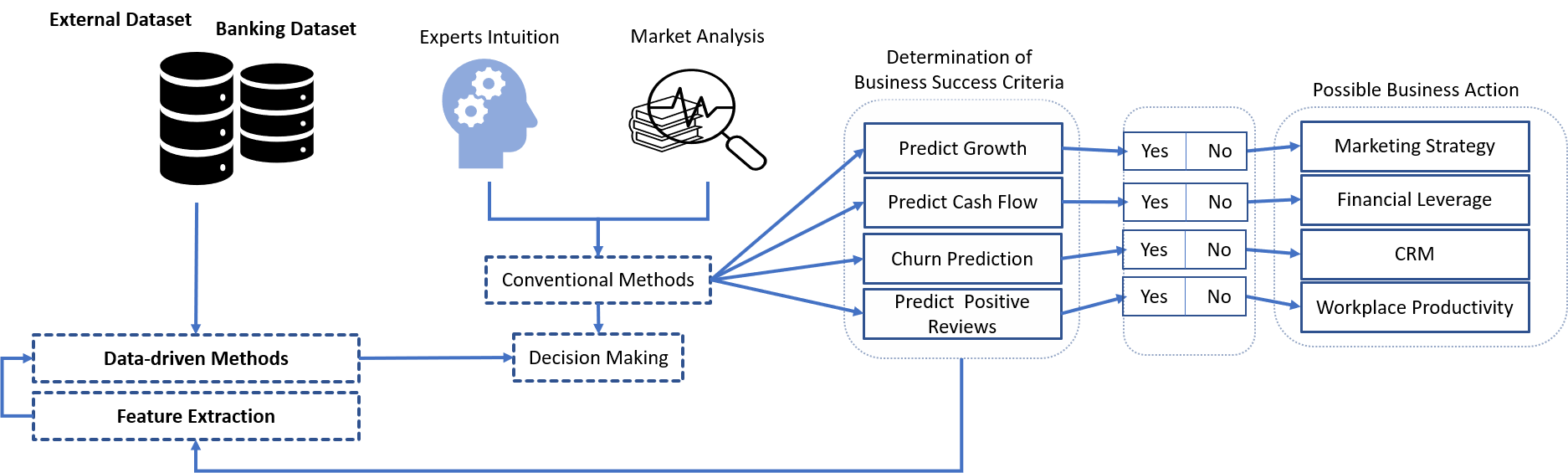}
        	\end{center}
        	\caption{Data-driven VS Conventional Decision Making.}
	        \label{Data-driven VS Conventional Decision Making2}
        \end{figure}

As already outlined in the introduction section, the purpose of this study is to produce a framework for banks and SMEs to make a data pipeline and create insights by predicting a label out of more than two groups. In contrast to previous studies with the central focus of independent prediction of each aspect of a business, business success breaks down the components~\cite{B-processanalytics-Beheshti} initially and aggregates later in this framework. This process will provide analysts to take benefits of the volume and the variety of big data as shown in Figure~\ref{Data-driven VS Conventional Decision Making2}.

Research suggests that we can see the changes in the semantic of the problem but no differences in machine learning mechanics, simply because our job is to separate the data points regardless of the meaning\cite{dataefficient}. Finding the line that best fits data and predicting the right label that a specific customer falls into which category can be done by logistic regression as a linear classifier and binary classification. However, In some decision-making processes in the banking industry, due to more than two target categories and more than two inputs, the dimension is greater than one and the vector equivalence that indicates a plane or hyper-plane is applicable with the well known $ \hat{y}=w^Tx+b $ equation.

While the traditional methods rely on the experts and intuitive feelings to extract the features, this study will focus on data-driven discovery to save invaluable time and cost. Multi-criteria decision-making by optimising one loss function can result in fault diagnosis during the feature extraction process. Taking advantage of \textit {Multi-task Learning} could provide an effective way to achieve knowledge transfer and extract the related features for each task. Table~\ref{StateofTheArt} illustrates the big picture of the diagnosis of ML\cite{dataefficient} in terms of how, when and what extent of knowledge is useful.

 \begin{table}[h!]
\caption{Overview of the Current State of decision making.}\label{StateofTheArt}
\centering
\begin{tabular}{ |p{4cm}|p{7cm}|  }

\hline
Sharing Knowledge & Purpose  \\
\hline
Transfer Learning~\cite{wei2017learning,TL_book1,TLsurvey,MTL-adaptiveknowledgetransfer,TL-MTL-CNN} & Aims to improve learning and minimize the amount of labeled samples
required in a target task by leveraging knowledge from the source task \\

Multi‑task learning~\cite{MTL-Classification-Task,MTLoverview-V1,MTLoverview2,MTLAttention,MTLNNArchitecture,Cross-Stitch-MTL} & If a TL method aims to improve the performance of the source task and target task
simultaneously, we are dealing with a Multi-task learning (MTL) problem \\
Meta‐learning~\cite{metaLearning,Metalearning2} & Improves the learning of a new task by using
meta-knowledge extracted across tasks  \\
 \hline
\end{tabular}
\end{table}

\subsubsection{Transfer Learning}

While a machine learning algorithm should have a generalisation feature, transfer learning indicates that the model has been used for training data, can predict future data. However, research~\cite{wei2017learning,CDCR} debates hand-picked features for the particular task affect performance using a training dataset with specific distribution and testing on different distributed data.

Transfer learning is taking the knowledge learned from one task and applying that knowledge to another task in the neural network\cite{TL_book1}. For example, the neural network learns to recognise objects like dogs on pairs of X and Y then apply part of that knowledge to the medical application to do a better job on radiology diagnosis. In other words, take the last output layer and weights feeding into that output layer of the trained neural network on image recognition and delete that. Then implement transfer learning by swap in a new data set and creating a new set of randomly initialised weights just for the last layer.

\subsubsection{Recurrent Neural Networks (RNNs)}

The theory relevant to attention mechanism and its connection to Multi-Task Learning will be examined in this sector~\cite{attent1}. Even though this study will not map a sequence-to-sequence model, the new idea of the attention algorithm is mostly replaced with encoder-decoder architecture for the conventional application of RNNs. Thus we will examine its impact to determine if there is a chance to improve the feature extraction for this study.
The human brain usually does not memories a paragraph or even a sentence when we want to translate a sentence from a different language. The current trend in the field of encoder and decoder in language model that can estimate a person's name in a given sentence to machine translation has been spread in many applications. There is a substantial body of literature~\cite{Encoder-Decoder,SequenceMain} of difficulties for NN to memorise the whole long sentence. The solution would be a kind of algorithm for the machine to perform like a human. We break down a long sentence to the smaller components, not necessarily to individual words, then translate that segment. After that, we will pick the next part, and finally, after developing the block, we will put them together and generate the decoded sentence.

\textbf{Attention-Based RNNs.}
The key idea of attention mechanism intuition was initially introduced by Bahdanau~\cite{AttentionmainArticle} in his paper for machine translation based on human visual attention. This new mechanism has overcome the problem of fixed-length vector in long sentences by the encoder using the neural networks that shares feature learned across the different position of the text.

By computing a set of attention weight to vectors, the model does identify and linearly weighted all the vectors. For example, in machine translation, the attention to words closer to the target word would be different for generating long sentences. In other words, when dealing with sequential data, the attention mechanism extracts the information from the whole sequence by attention to the appropriate data point of the input to predict the next output element.
the attention mechanism assumes a sequential dataset that each data point can be seen as a vector
$V_{i}$
and while we have ($V_1$ to $V_n$) with the fixed dimension, the length of the sequence is unknown. The ultimate goal is to train the model to learn the optimal weight when we want to perform a linearly weighted sum of the vectors.

Next, we will review the multi-task learning model to apply for common feature learning from the same input data to learn multiple tasks simultaneously~\cite{MTLAttention}. Therefore, as we will discuss in the methodology section, we would like to form a global feature pool, learn task-specific features, and train the model to learn the optimal weight of each feature.

\subsubsection{Multi-task Learning}

Human learns based on experiences. When we want to make a general decision, we do not consciously label every underlying component as true or false. An example could be the way we apply our soft skills in a different situation. Once we learn how to get along with one person, our memory guides us to find new friends, improve our relationship, manage some conflicts in a better way and generally have an impact on day-to-day interactions with the rest of the world. While the final output is uplifting the quality of life, the question is how our brain maps the features of every situation and transfers those skill sets from one situation to another. One of the essential features of human cognition is memory, and another is attention. In this section, we will explore these two features and their impacts in the ML field.

\begin{figure}[t]
        	\begin{center}
        	\includegraphics[width=0.9\linewidth]{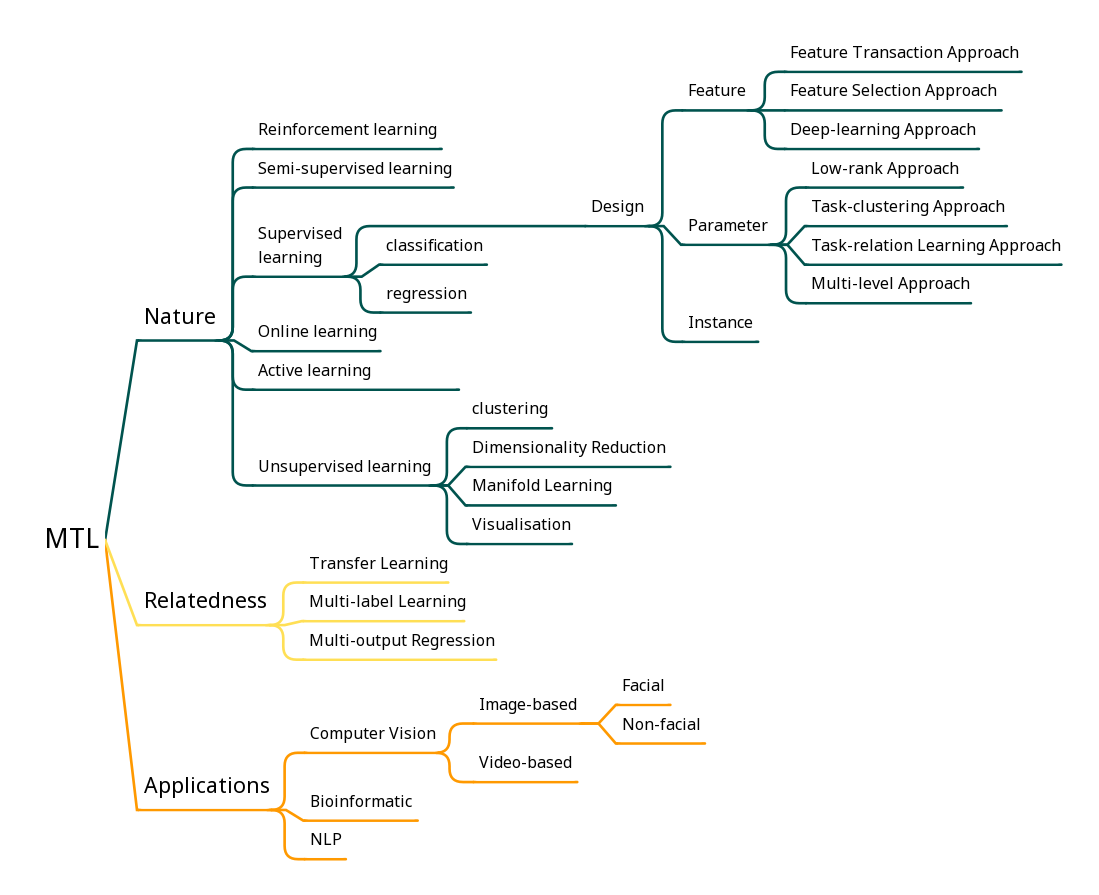}
        	\end{center}
        	\caption{Multi-task learning Direction in Machine Learning.}
	        \label{MTL overview}
        \end{figure}

As we will see in Figure~\ref{fig:Toatal Data Set Half Size},
There are shared featured available for several related tasks. Therefore, finding the feature selection approach~\cite{MTLoverview2} with the ability to represent the most associated subset of the original feature for different tasks would be our main concern.
Yu Zhang and Qiang Yang have tried to categorised MTL based on the nature of tasks, relatedness to other areas of machine learning and applications in their survey~\cite{MTLoverview2}, which is depicted in Figure~\ref{MTL overview}, and clearly illustrates the majority of accomplished works are in the supervised learning field.
Table~\ref{Table MTL} shows the category and applications of MTL from the theoretical points and their applications when the goal of the ML application is optimising more than one loss function.

There are two structures of the MTL model ~\cite{MTLoverview-V1} to operate in deep neural networks that aims to improve the learning when all or a subset of tasks are related. The most common method uses hard parameter sharing to share the low-level hidden layer in the neural network between all tasks and split the layer at a higher level or task-specific layer. This mechanism is famous because we can avoid overfitting since the simultaneous learning process happens at the early stages. Also, it helps to avoid the negative transfer. The inappropriate definition of a task relationship may cause the risk of negative transfer. Even though This method needs to define task relationships beforehand, the underlying assumption of the closely related task in MTL may not meet for each task associated to all of the available tasks and thus exploit the performance.

Based on the research problem, the lack of application of MTL in Multiple class classification and relatedness to another field of ML as shown in Figure~\ref{MTL overview}, we will employ the soft parameter sharing MTL technique to solve the feature selection and prediction of success criteria for small and medium-sized businesses regarding data collected from external large data sources and bank. It is crucial to mention lots of features are shared to predict different criteria. The soft parameter sharing approach is a mechanism designed to share parameters among tasks while each task has its parameter.

\setlength{\arrayrulewidth}{0.2mm}
\setlength{\tabcolsep}{18pt}
\renewcommand{\arraystretch}{1.5}
\begin{table}[t]
\caption{Categories of MTL and their application Classification.}\label{Table MTL}
\centering
\begin{tabular}{ |p{2cm}|p{2cm}|p{3cm}|p{3cm}|  }
\hline
Categories of MTL& Problem& Relatedness to Other Area of ML& Applications  \\
\hline
Supervised Learning & Classification & Transfer Learning & predict labels for unseen data~\cite{MTL-Classification-Task}\\

 & Regression & Transfer Learning & problems encountered in the high dimensional vectors~\cite{MTL-Regression-Task}\\
\hline
\end{tabular}
\end{table}

\subsubsection{Attention Mechanism}

Massive improvements in encoder-decoder architecture in RNNs results in new architectures that can get single data points and turn a static vector into sequential processing.

\begin{figure}[h!]
        	\begin{center}
        	\includegraphics[width=1\linewidth]{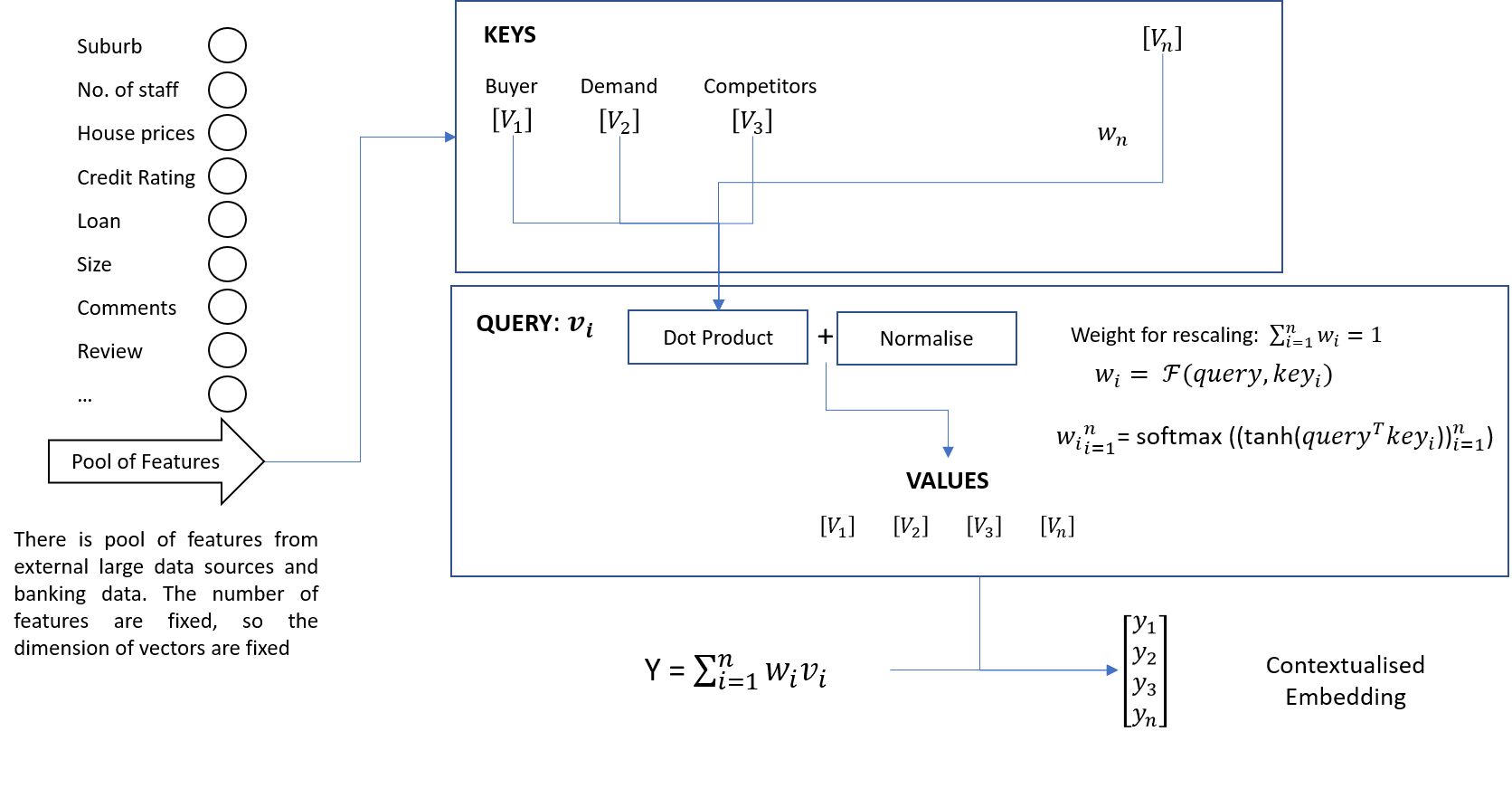}
        	\end{center}
        	\caption{The Architecture of  Attention Mechanism ~\cite{Attentionisallyouneed}.}
	        \label{Attention}
        \end{figure}

In a competitive business environment deliver a reliable recommendation to the customers is critical. Given the end-to-end neural architecture with a domain expert's help to extract the feature still plays a dominant role in the big data environment~\cite{MTLNNArchitecture}, While in real-world use cases we need an algorithm with the intention of not being a single task learner in a single environment.
Very similar to NN's background concepts, which has been derived from human brain activities, as shown in Figure~\ref{Attention},
there is a method to learn a common feature set for different tasks based on human visual attention.
In 2017 the paper~\cite{Attentionisallyouneed} published by a group of researchers who were mainly interested in natural language processing. Since then, this new mechanism has almost applied on top of any classification task. This mechanism's main application came to help recurrent neural networks when there is sequential data, and each data point is a vector. So, whenever we have a notion of time, this mechanism will increase the performance of RNNs. For example, in translation application, while applying RNNs with encoder-decoder architecture for long sequence, the decoder only access the most recent sentence, and extracting information from the whole sequence is almost impossible.

Being inspired by the ability of the human brain to focus on one thing and ignore other things, there is a challenge in machine learning architecture to learn from a different domain without the need for transfer learning just by changing attention to features. As shown in  Figure~\ref{fig:Method}, if we can train a neural network for demand prediction that its inputs is business extracted features, But also, predict the next task as a potential buyer by taking the previous NN and transfer what is learned to this new task by deleting the last output and associated weight feeding to that last output layer. But here, we have the same input for both tasks. The only difference is the weight of those parameters for each task.

Even though customer action histories are sequential, we still need to justify the attention mechanism for this experiment since the features for our criteria do not look sequential. But the goal of this study is to generalise the model to share feature learned across different tasks, then if we can assume:

\begin{itemize}
\item Each data point is a vector with fixed dimension
\item The length of the sequence is unknown
\end{itemize}

we are free to do not provide any definition for tasks before the learning process. Like methods such as client2vec, customer2vec, item2vec, and word2vec, this structure is helpful to learn similarities based on shared features.


\section{Methodology}\label{Chapter 3}

\subsection{Preliminaries}

\begin{figure}[h!]
        	\begin{center}
        	\includegraphics[width=0.9\linewidth]{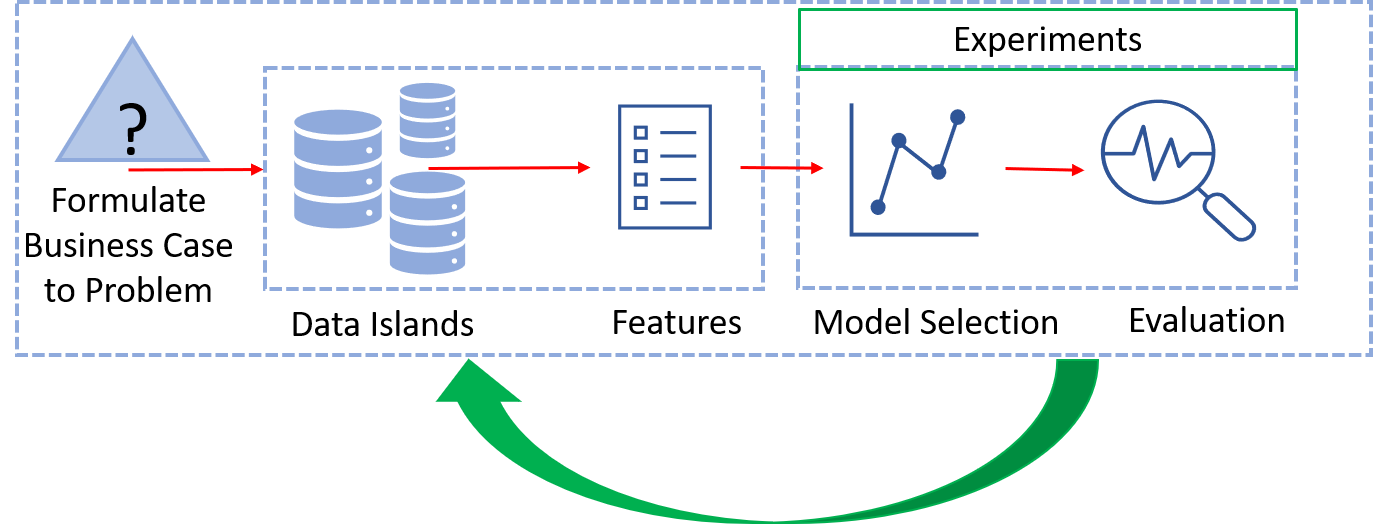}
        	\end{center}
        	\caption{A general view of formulating the big data problem in establishing a new business.}
            \label{fig:model1}
        \end{figure}

This section will explain the systematic procedures to answer the research question in a logical and orderly way. This paper proposes a novel data product in the banking industry for potential SME customers who want to establish a new business.
Figure~\ref{fig:model1} illustrates a general view of formulating the big data problem in establishing a new business.
%
The method framework is described at Algorithm \ref{algorithm_2}.
\begin{algorithm}
 Building Blocks;\\
 \KwData {
  1)Identifying Data Islands, 2) Feature extraction
 }
 Multiple-class Classification;\\
 Multi-task Learning;\\
 Prediction of Successful SMEs using Different Classifiers ;\\
 \KwResult{ Deliver Insight to a Prospective Customer}
 \caption{Successful SME Customer Prediction Framework}
 \label{algorithm_2}
\end{algorithm}
This process requires good collaboration between humans and technology, developing a pipeline of available data from external large data sets and banking data set, forming a big data platform, extract features, linking features to business criteria, predicting, and creating insights.

A careful analysis of the stated problem reveals a multi-criteria decision-making challenge with several inputs and more than two outputs. Experts used to draw a two-dimensional scatter plot with two inputs and draw a line between the two classes, successful or unsuccessful. However, for linear classification in higher dimensions, separating the data is not a line but a plane or a hyper-plane. To formulate this problem and explore the predictive model for regression and classification for the same input data, the Multi-Task Learning (MTL) problem's generalisation methods have been employed based on published articles~\cite{Zhao2020,Banklocation}. The early models proposed a single neural network AI technology. Some models are based on artificial neural networks guided by genetic algorithms to identify, target, and profile most likely consumers. Later, multi-task learning models with many attributes tackle the problem with predefined task relationships and guide the learning process.

\subsection{Building Blocks}

Scholars have argued that separating the data points into different categories by defining line, plane, or hyper-plane has been constructed as a black box~\cite{Blackboxmodels}.
This investigation approach covers an in-depth analysis of the previous classifier publication and state-of-the-art that Zhao and his colleagues~\cite{Zhao2020} have published.
It is reasonable to examine the black box decision systems approach for this successful business prediction model with a simple logistic regression function to support later classifiers. Given a problem definition and having labelled data, we are facing supervised learning. We want to split the data into categories (successful or unsuccessful). In other words, to predict whether the potential business has a chance to survive more than five years or not. From the input data, we want to predict the likelihood of demand by people who are potential buyers in the specific area. Also, from the same input data, the prediction of turning the buyers to the customer is another concern. As such, the competitors can impact the newly established business, which is another matter that needs to be predicted. In the next section, we will create multiple binary classifiers, neural network and multi-task learning model to compare the results. But let's do not overlook the process of learning from one task and transfer to another task. Instead, trying to have a NN, get several things done simultaneously that each task will help another task.

\subsection{Multiple-Class Classification}\label{multiple section}

\begin{figure}[h!]
        	\begin{center}
        	\includegraphics[width=1\linewidth]{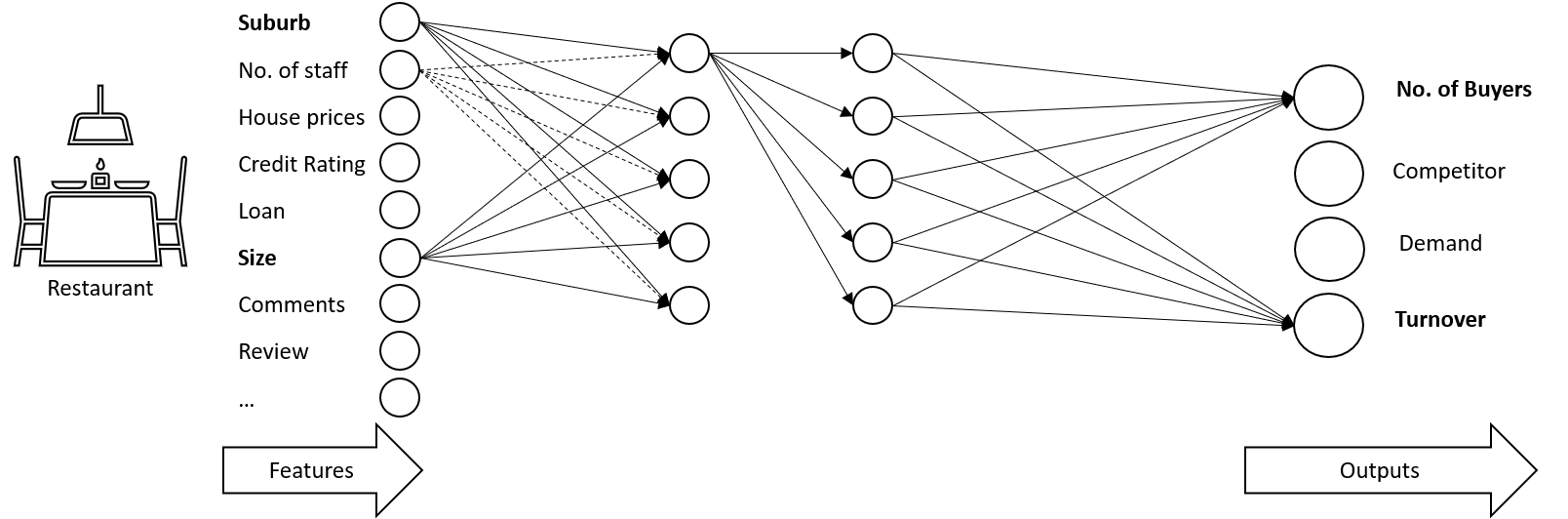}
        	\end{center}
        	\caption{Multiple-class classification of a restaurant.}
            \label{fig:Multiple-class classification}
        \end{figure}

To solve the problem, we first need to define the output we want to predict. In this section, as an example, a restaurant and its potential components are considered. As shown in the Figure~\ref{fig:Multiple-class classification}, from the different types of features, we can identify multiple classes. In other words, for only one data point, there are various tags available. Some of them are continuous, and some binary classes, while at this stage of analysis, the types and number of features of the data point is not our concern.  For example, while store dimensions within a neighbourhood can be the primary determinant for the number of buyers in one suburb, another suburb's socioeconomic characteristic can positively or negatively impact that.

Consequently, it would seem better to define a multi-class problem since predicting an active business for more than five years as an ultimate goal have multiple neurons. On the other hand, for this non-exclusive multi-class situation, the output layer has multiple neurons whose loss function can be optimised separately. In this challenge, the knowledge comes from the same domain~\cite{TLsurvey}, not a related domain similar to what we usually see in other disciplines of ML applications like NLP or image processing for medical purposes. So we have to find another way to improve the performance on each task.

\subsection{Multi-task Learning}\label{MTL}

        \begin{figure}[h!]
        	\begin{center}
        	\includegraphics[width=1\linewidth]{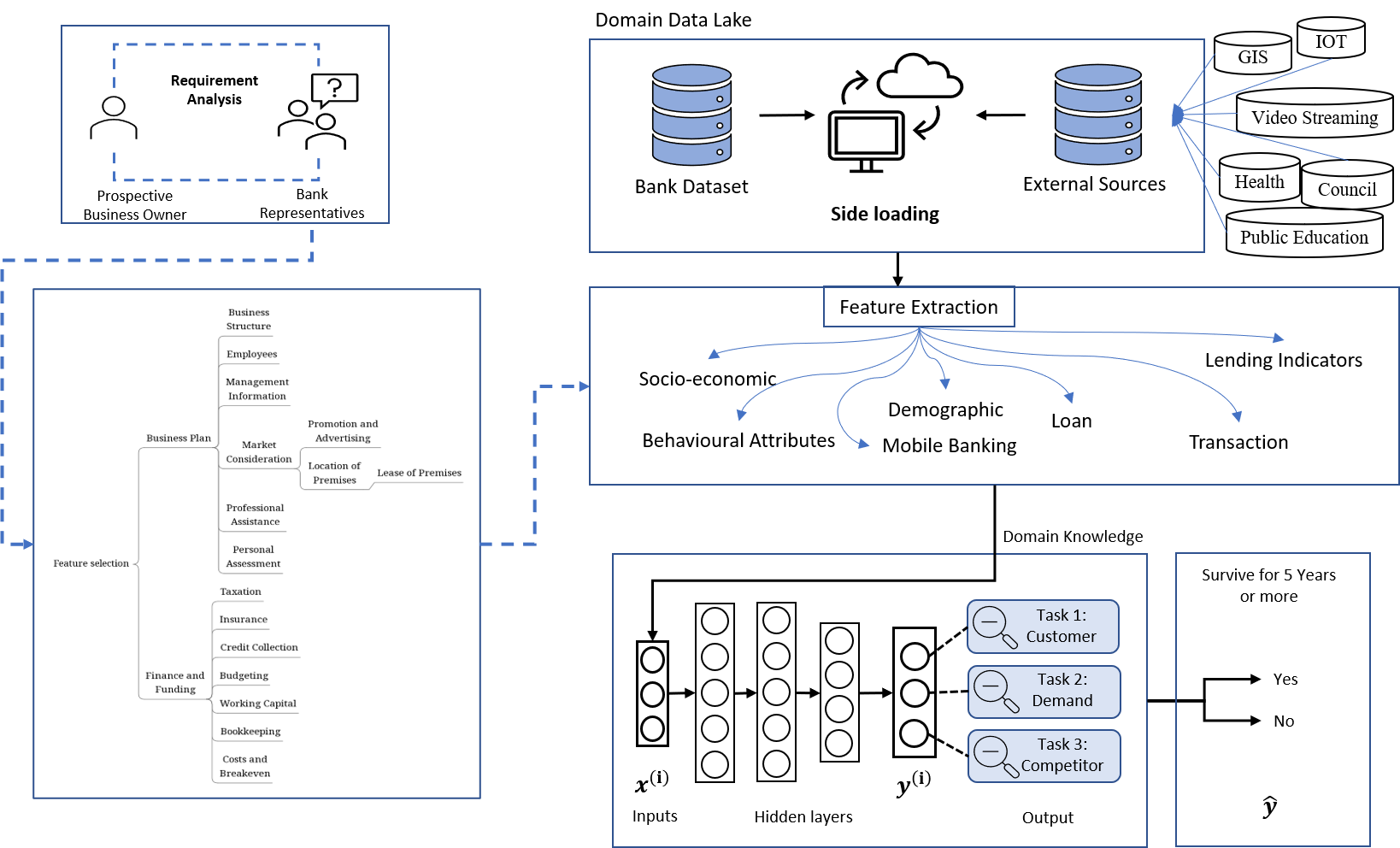}
        	\end{center}
        	\caption{Black Box Architecture.}
            \label{fig:Method}
        \end{figure}

Lacking a large number of labelled data is considered a bottleneck in many deep learning NN algorithms. For example, in a proposed framework, the algorithm has to learn from the very limited number of instances. Based on the report published in 2019, the total number of Australian small businesses recorded as 2.3 million~\cite{SB_report} that is divided into 18 industry sector. In a very probable situation, the sample size for each sector for a given bank would never go over several thousand. Therefore the first problem to tackle is to find a way to increase the sample size for training the model.
As we see in Figure~\ref{fig:Method}, we have three tasks, and the amount of data for each task are similar. If we wanted to train only one task in isolation, the number of examples would be much smaller, similar to what we usually do in transfer learning. Although here we can take advantage of other tasks to provide some knowledge for each other.
In CNNs and RNNs, having a problem-independent feature extraction can overcome this problem. In this study, we are not confronting unlabeled data but taking advantage of the developed generic methods to learn the most associated feature for each task is applied to predict the successful new established business.

Let us look at the different things we want to predict in the matrices form;
\begin{equation}
y =    \begin{bmatrix}
        Potential Buyers\\
        Demand\\
        Competitors
      \end{bmatrix}
        \label{first matrices}
\end{equation}

To have a better understanding of problem formulation, consider a
customer ($Customer A$) who has a medium level of buyer with high-level demand and a low level of competitor. If this customer is an input for our model, then instead of having one label ${y^{(i)}}$, there are three labels. And, if we plan to predict other criteria, ${y^{(i)}}$ can have more dimension. So, ${y^{(i)}}$ is $(3*1)$ vector.
Therefore, the training set label as a whole shown in Formula~\ref{Whole matrices} that is in our study a $(3*m)$ vector.

\begin{equation}
Y=
\begin{bmatrix}
{y^{(1)}} =
    \begin{bmatrix}
        Potential Buyers\\
        Demand\\
        Competitors
    \end{bmatrix},& {y^{(2)}} =
                    \begin{bmatrix}
                             Potential Buyers\\
                             Demand\\
                             Competitors
                    \end{bmatrix}&, ..., &{y^{(m)}} \\

\end{bmatrix}
\label{Whole matrices}
\end{equation}

Our goal is to optimise three loss functions simultaneously to helping business banking customers with insights into whether their new business will survive more than five years or not.in other words, train a neural network, for input${x}$ and output is 3-dimensional value for ${\hat{y}}$.
To train this NN, the loss function for given predicted output which is ${\hat{y}^{(i)}}$, defined as the average usual logistic loss of individual predictions of $({\hat{y}^{(i)}}_{j},{{y}^{(i)}_{j}})$:

\begin{equation}\label{equation3formula}
{\frac{-1}{m}} {\sum_{i=1}^{\infty}}{\sum_{j=1}^{3}}
({{y}^{(i)}_{j}} { log } ({\hat{y}^{(i)}}_{j})+ (1-{{y}^{(i)}_{j}}) { log } (1- {\hat{y}^{(i)}}_{j}))
\end{equation}

In Subsection~\ref{multiple section}, we discussed the necessity of simultaneous multiple class classification, and Equation~\ref{equation3formula}, illustrate the MTL structure to generalise the prediction method.
 Still, there is a challenge if we want to include more criteria or tasks in this model because some of the tasks might be unrelated or loosely related ~\cite{MTLAttention,Zhao2020}. To design this framework, we will use soft parameter sharing plus a mechanism to capture the relationship without relying on a predefined task relationship to avoid the stated problem.

\subsection{SME Prediction}

There are academic literature in formulating the determinant factors as a general linear model to predict the business success
~\cite{impactofsocio-economicfactors} as shown in Equation~\ref{linear equation}:

\begin{equation}\label{linear equation}
Y=\beta_0 + \beta_iX_i + e_i
\end{equation}

where $X_i$ can be made up of variety of independent variables and $e_i$ as a error term.
However, the limitation of linear analysis and properties of error term ~\cite{Errorterm} with assumption of no systematic pattern, has directed us to start our experiment with ML algorithms.

\subsubsection{Logistic regression}

In terms of machine learning literature, logistic regression provides a good baseline model. While it is well interpretable, fast, established by statisticians for many years, we do not need to perform lots of tunning like deep neural networks. Meanwhile, in high-level analysis, logistic regression as a neuron and the combination of these neurons will form neural networks.
To transform a number to the category, first, encode two categories as 0 and 1 and pass:
\[\hat{y}=w^Tx+b\]
 into the sigmoid function. The result always
 returns a number between 0 and 1. Then interpret this as the:
\[P(y= 1 | x)=\sigma(w^Tx+b)\]
The outcome only gets exactly 0 or 1 when the input is negative or positive infinity due to the asymptotes on either side of the sigmoid. The sigmoid takes in any actual number, and if that number is more significant than zero, it returns a number greater than 0.5. If the input is less than zero, then it returns a number less than 0.5. And typically, we use 0.5 as a threshold, so if
$P_y = 1$
given x is more significant than 0.5, we predict 1; otherwise, we predict zero.

In the next section, we will present a motivating scenario and the result validation of the experiment obtained from the proposed method compared with the baseline. We will discusses how this approach could help the enhancement of the effectiveness in predicting successful businesses.


\section{Experiment and Evaluation}\label{Chapter 4}

This section aims to present the result validation of the experiment obtained from the proposed method compared with the baseline and discusses how this approach could help the enhancement of the effectiveness in predicting successful businesses.

\subsection{Banking Dataset and External Sources}

\begin{figure}[h!]
        	\begin{center}
        	\includegraphics[width=1\linewidth]{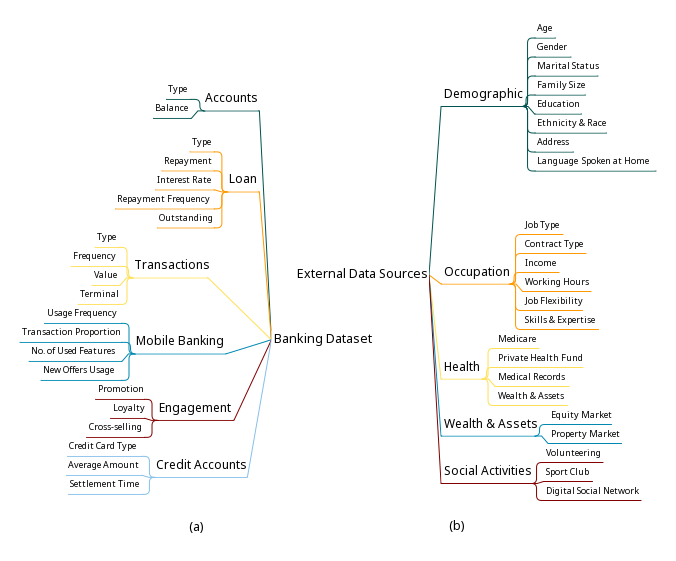}
        	\end{center}
        	\caption{A snapshop of the data about the customers, products, and accounting categories~(A) and a set of publicly available datasets~(B).}
            \label{fig:Toatal Data Set Half Size}
        \end{figure}

The techniques employed to solve the problem has started with available relevant data sources to address the business problem as a central component of the analysis. from the banking side, Tables that contain the data about the customers, products, and accounting categories are employed as shown in Figure~\ref{fig:Toatal Data Set Half Size}a,
the next step is to ingest relevant data from external large dataset. Publicly available data as shown in Figure~\ref{fig:Toatal Data Set Half Size}b from government agencies like Australian Research Council, Australian Bureau of Statistics, Tourism Research Australia, and council's data have been used to form the data profiling.

For each data points in our dataset,  there are three tasks in numeric value that is labeled in Low, Medium and High classes; potential buyers, demand products or services and competitors, as illustrated in Table~\ref{task}. The instance's measurement process involves direct and indirect approaches. The number of businesses in the same category that competes in the identical postcode is collected from the council dataset
I.e., people counter using a cellular network or wifi connection is a relatively easy task because intelligent devices like the mobile phone, GPS devices and other forms of IoT devices are vastly using by customers and companies these days. The footfall analysis is not a part of this experiment, and thus the number of potential buyers has been collected from third party companies.
Demand is measured by a closed sale deal for a product and calculated by the number of transactions for each data point. Reserve Bank of Australia has conducted a consumer payments survey~\cite{paymentmethods} that indicates 63\% of Australian consumers prefer electronic payment methods even though for lower-value payments, cash is preferred.

\setlength{\arrayrulewidth}{0.2mm}
\setlength{\tabcolsep}{18pt}
\renewcommand{\arraystretch}{1}
    \begin{table}[!t]
        \caption{An Overview of Tasks and Measurements.}\label{task}
        \centering
        \begin{tabular}{
       c c c c c }

            \hline
                Instances&  Buyers-Label& Demand-Label& Competitors-Label\\
            \hline\hline
                A &  10000-M & 361-H & 50-L\\
                B &  12500-L & 139-H & 135-L \\
                C &  6000-H & 248-H & 350-H \\
                ... & ... & ...& ... \\
                Z & 21000-M & 118-H & 43-L\\
            \hline
        \end{tabular}
    \end{table}

\subsection{Experiment Progress}

There are academic literature in formulating the determinant factors as a general linear model to predict the business success
~\cite{impactofsocio-economicfactors} as shown in Equation \ref{linear equation}:

\begin{equation}\label{linear equation}
Y=\beta_0 + \beta_iX_i + e_i
\end{equation}

where $X_i$ can be made up of variety of independent variables and $e_i$ as a error term.
However, the limitation of linear analysis and properties of error term ~\cite{Errorterm} with assumption of no systematic pattern, has directed us to start our experiment with ML algorithms.

\subsubsection{Logistic Regression}

In terms of machine learning literature, logistic regression provides a good baseline model. While it is well interpretable, fast, established by statisticians for many years, we do not need to perform lots of tunning like deep neural networks. Meanwhile, in high-level analysis, logistic regression as a neuron and the combination of these neurons will form neural networks.

To transform a number to the category, first, encode two categories as 0 and 1 and pass:
\[\hat{y}=w^Tx+b\]
 into the sigmoid function. The result always gives back a number between 0 and 1. Then interpret this as the:
\[P(y= 1 | x)=\sigma(w^Tx+b)\]
The outcome only gets exactly 0 or 1 when the input is negative or positive infinity due to the asymptotes on either side of the sigmoid. The sigmoid takes in any actual number, and if that number is more significant than zero, it returns a number greater than 0.5. If the input is less than zero, then it returns a number less than 0.5. And typically, we use 0.5 as a threshold, so if P of y equals 1, given x is more significant than 0.5, we predict 1; otherwise, we predict zero.

\subsubsection{Multi-class Classification Approach}

\begin{figure}[t]
        	\begin{center}
        	\includegraphics[width=0.8\linewidth]{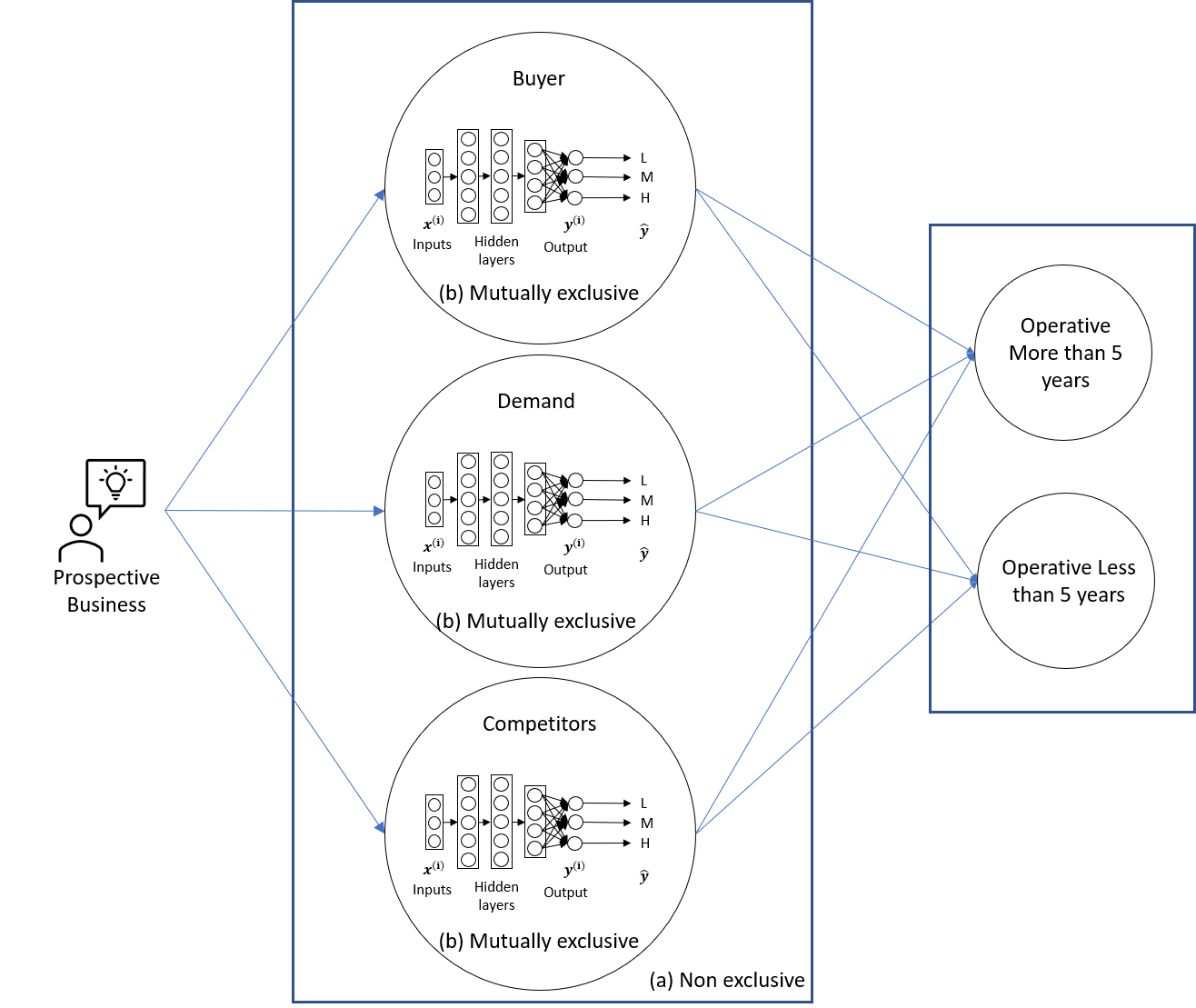}
        	\end{center}
        	\caption{Types of Multi-class Classification}
            \label{fig:Multi-class}
        \end{figure}

As illustrated in Table~\ref{task}, we realise that each of those three tasks has three different possible options. Each instance can have only one class (active more OR less than five years) assigned to it. We will break down the problem into two parts, as shown in Figure~\ref{fig:Multi-class}, while the second part deals with non-exclusive classes because each data point can have multiple tags, the first part can only have one output node per classes, or it is a mutually exclusive class. For example, data point A can be categorised as a business with an average number of potential buyers with high demand and low competition level. Still, competitors themselves can choose only low, medium or high.

Using one-hot encoding will help solve the issue to assign a label to each instance, and the next step is to choose a correct activation function for each. The sigmoid function for the last output layer is selected for the non-exclusive part with correctly organised data. Therefore, the likelihood of having demand, buyers and competitors is calculated.
For the first part, since we have three different events, the softmax activation function would be the best choice. The calculation would be the probabilities of each target class over all three possible targets.


\subsubsection{Training the Dataset }

P of Y given X represents a full probability distribution over all individual values inside the Y matrix given the X matrix. Since all of the neural network weights are random, output probabilities (predictions) are belongs to each class is also random. For the training part,  the  fit(X,Y) function is used from Scikit-learn libaray.
\[P(y | x)=softmax(V^Tf(w^Tx))\]
So the way we have applied to train the logistic regression and neural network are the same.
The following steps are considered:

\begin{itemize}
    \item Define cost or loss function, which is the error between prediction and target. For this study, Mean Squared Error for predicting Demand, Number of buyers and competitors.
    \item Minimise the cost (L) with respect to W
\end{itemize}



\subsection{Evaluation Metrics}
\begin{figure}[t]
        	\begin{center}
        	\includegraphics[width=0.8\linewidth]{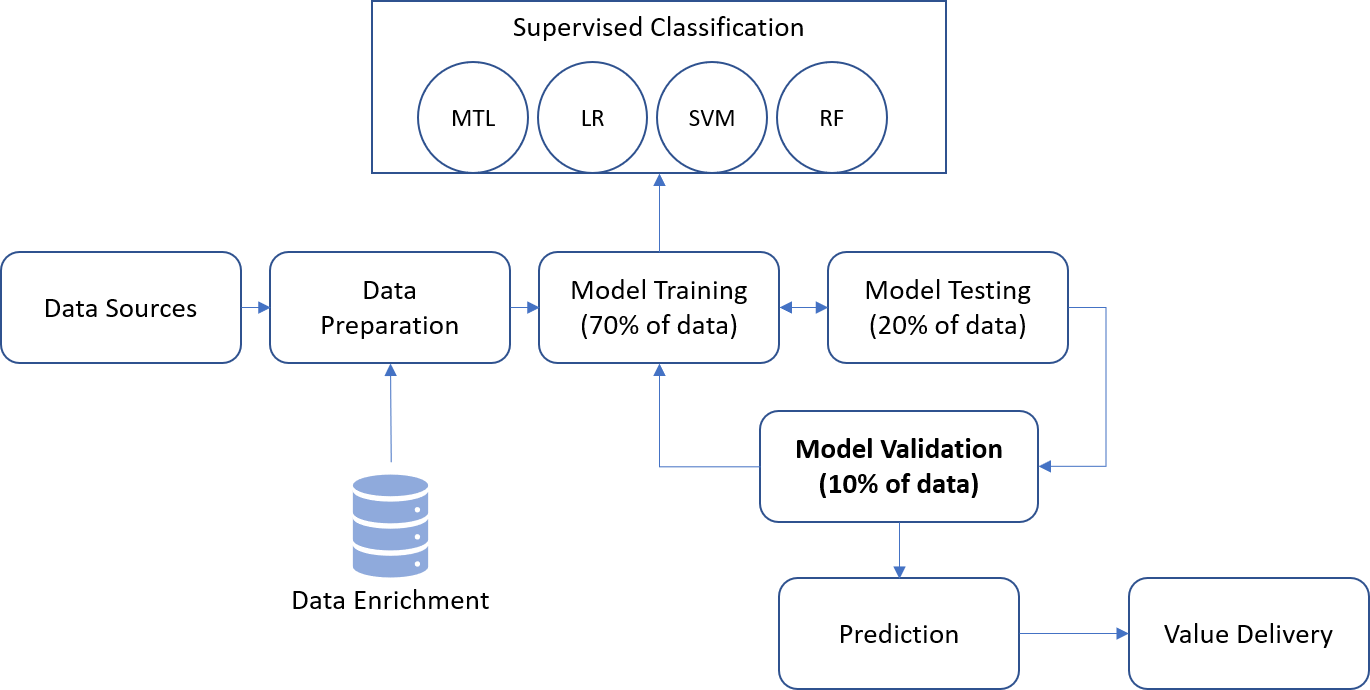}
        	\end{center}
        	\caption{Validation Importance in Conventional ML Architecture.}
            \label{fig:Evaluation}
        \end{figure}

Even though today's ML practices have found lots of success in many applications such as NLP, speech recognition, and computer vision, insights from one application are not easily transferable to another area. As shown in Figure~\ref{fig:Evaluation}, we have followed the iterative analysis in evaluation of different set up on training, development, and test data for this project. Also, for the neural network part, we are concerned about the number of layers, the number of hidden units for each layer, the learning rate, and activation functions for different layers.

\begin{wrapfigure}{t}{0.5\textwidth}
        	\includegraphics[width=0.5\textwidth]{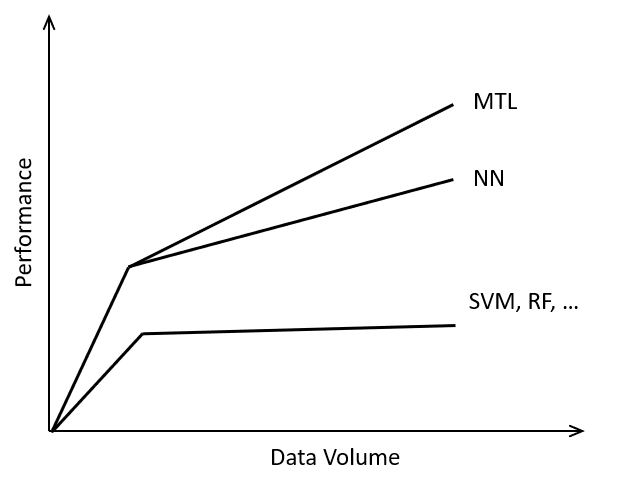}
        	\caption{Expectation of Applying Different Classifiers }
            \label{fig:Classifiers}
        \end{wrapfigure}

According to Andrew Ng, the amount of data for a task and the  performance of a learning algorithm or accuracy usually follow the pattern is shown at Figure~\ref{fig:Classifiers}.

The process of improving machine learning efficiency may include collecting more data, reconsider the diversity of training data, or trying a different optimization algorithm~\cite{EvalautionOreilly1}. The proposed system has to make sure that performance on the training set can pass the acceptability assessment. Finally, doing well on the test set on the cost function results in the system performing in the real world.

The selection of association ML algorithm to the business requirements for this project guide us in four different supervised learning models because of the nature of structured data and the K-fold cross-validation scores to decide on hyperparameter tuning. Our goal is to outperform accuracy in prediction classifiers~\cite{EvaluationOreilly2} such as SVM and Random Forest by neural networks for a single task and, in particular, form MTL and share representation between tasks. Before predicting, we have to validate the model over the data that the model has not seen earlier during the training process. This step enables us to generalise the model for the upcoming dataset.

In terms of the binary classification problem, we want to predict whether a business can survive more than five years or not. From the confusion matrix point of view, four possible alternatives are plausible:
\begin{itemize}
\setlength\itemsep{-0.5em}
  \item True Positive: predicted unsuccessful business and is unsuccessful
  \item False Positive: predicted unsuccessful business while is successful
  \item True Negative: predicted unsuccessful business and is unsuccessful
  \item False Negative: predicted successful business while is unsuccessful
\end{itemize}
From the above outcomes, we presume the false negative is the most destructive prediction we can get. To reduce the risk of performance of our ML model, we consider Specificity as ML metrics which is the percentage of unsuccessful cases out of the aggregate actual unsuccessful datapoints instead of recall that stress on the successful instances.

\setlength{\arrayrulewidth}{0.2mm}
\setlength{\tabcolsep}{18pt}
\renewcommand{\arraystretch}{1}
\begin{table}[t]
\caption{Metrics for Evaluating the Performance of Approaches.}\label{Table evaluation}
\centering
\begin{tabular}{ |p{3cm}||p{1.5cm}|p{1.5cm}|p{1.5cm}| }
 \hline
 \multicolumn{4}{|c|}{Result of the performance in Terms of Root Mean Square Error} \\
 \hline
 $\hat{y}_i$  & Demand& Number of Buyers&Competitors\\
 \hline
 Random Forest   & 1111.6   &2513& 173.2\\
 Support Vector Machine   & 1985.5   &2113.6& 104.3\\
 ANN   & 1843.2   &2364.2& 149.3\\
 Multi-task learning   & 1680.3   &1760.3& 236.2\\

  \hline
\end{tabular}
\end{table}

Apart from the established method to measure the performance of the ML model, we can see the novel methodologies to evaluate the ML solutions. For example, according to G. Barash et al., a gap between existing ML solutions and associated business requirements ~\cite{Ml_solutionB_requirement} is modelled as different combinations of the input configuration. And therefore, it can be evaluated by classical combinatorial testing technique to detect inadequately tested business areas of ML solution. They believe their proposed model based on testing on three real case studies will improve the ML performance over data slices.


\section{Conclusion and Future Work}\label{Chapter 5}

\subsection{Conclusion}

This study proposed a new data product in the banking industry for a prosperous small and medium-sized business that has not been established yet by delivering business insights to potential customers using big data. Establishing a new business need to make a tactical business decision based on discovering purchasing behaviour, consumer buying signals, demographic and socioeconomic attributes of different locations. Due to economies of scale, the bank can afford to obtain external data and multidimensional analysis. And thus can described why and when something went well for current business customers employing past transactional data.

The proposed framework considers ingestion of external data sources to banking big data platforms, creating domain data lake, selecting and extracting features, predicting different criteria of a successful business, and delivering insights. The experiment is conducted on a dummy dataset created from Kaggle and the UCI machine learning repository. At this stage, the attributes can be found in government, council, public education and health data sources. Data types are chosen to multivariate since both regression and classification are applied in this experiment.

Based on the Australian small business and family enterprise ombudsman report, lasting more than five years is selected as a proxy for successful businesses. The predicted criteria are demand, customers, competitors. Since the number of instances for every business category is limited and the challenge is multi­ criteria decision making, the method to tackle this problem is Multi-task learning with an attention mechanism. Results showed that developing the regression and classification predictive model on the same input data is a more effective way in the case of limited instances. And employing the attention mechanism to set the task relationship instead of a predefined task, we can improve the result by using multi-task learning instead of transfer learning.

\subsection{Area of Future Work}

Very similar to the human brain, data products that aim to improve the intelligent domain knowledge~\cite{Data-driveninnovation} for the decision-making process have to be able to query heterogeneous data sources. Even though, retrieval is a necessary process but not sufficient. We need other mechanisms to direct the attention of those recovered information into the categorised and valuable form for the user request. So the general framework with the ability to focus on the related component is vital in the highly competitive world of big data. Intelligent sumamrization approaches~\cite{SamSum1,iSheets,iCOP,iProcess} would also of high interest in analyzing banking data.
Other future applications in analyzing banking data for a successful customer engagement may include trust~\cite{Moh1,Moh2,Moh3,Moh4,Moh5}, recommender systems~\cite{shah1,Shah2,shah3,shah4}, and cognitive science~\cite{p2vec,khatam,vah1}.

Modelling the banking data using graph data models~\cite{graph1,graph2,graph3,graph4,dream} will facilitate the analytics and insight discovery in our future work. In this line of work, we plan to use storytelling approaches~\cite{story1,story2} to facilitate the discovery of facts from the data lakes~\cite{CoreDB} and knowledge lakes~\cite{CoreKG}. We will use crowdsourcing techniques~\cite{kushcrowd,crowd1,crowd2,crowd3} to mimic the knowledge of domain experts and enhance the feature engineering step.

The remarkable success in applications that applied multiple supervisory tasks in classification, detection and segmentation by sharing representation between tasks indicates that there is no ultimate answer for the structure of the best shared and task-specific representation in most application. To design the proper structure, \textit{Cross-Stitch Networks} for Multi-task Learning~\cite{Cross-Stitch-MTL} proposes linear combinations architecture to learn the optimal linear combinations for a given set of tasks that can be trained end-to-end. Future research will reveal whether we can find the simultaneous evaluation method for decision-making, feature selection, multiple tasks, domain knowledge, and source of datasets in the big-data arena.

\section*{Acknowledgements}
- We Acknowledge the AI-enabled Processes (AIP\footnote{https://aip-research-center.github.io/}) Research Centre and Tata Consultancy Services (TCS) for funding this research.

\bibliographystyle{abbrv}
\bibliography{ms}

\end{document}